\documentclass[10pt,twocolumn,letterpaper]{article}

\usepackage{cvpr}              %

\usepackage{graphicx}
\usepackage{amsmath}
\usepackage{amssymb}
\usepackage{booktabs}
\usepackage{multirow}

\usepackage[pagebackref,breaklinks,colorlinks]{hyperref}

\usepackage[capitalize]{cleveref}
\crefname{section}{Sec.}{Secs.}
\Crefname{section}{Section}{Sections}
\Crefname{table}{Table}{Tables}
\crefname{table}{Tab.}{Tabs.}

\def\cvprPaperID{3} %
\def\confName{CVPR}
\def\confYear{2022}

\usepackage{xcolor}
\usepackage{xparse}

\usepackage[normalem]{ulem}
\usepackage{ifthen}
\newboolean{final}
\setboolean{final}{false}

\newcommand{\FontFormatter}[2]{%
    \textcolor[rgb]{#1}{\textbf{#2}}%
}

\NewDocumentCommand{\TextFormatter}{ommm}{%
    \FontFormatter{#4}{%
        \IfNoValueTF{#1}{%
            $\scriptstyle \overset{#3}{++}$ #2%
        }{%
            \sout{#1} $\scriptstyle \xrightarrow{#3}$ #2%
        }%
    }%
}

\NewDocumentCommand{\TextFormatterFinal}{ommm}{#2}

\NewDocumentCommand{\ms}{om}{%
    \TextFormatter[#1]{#2}{Mustafa}{0.5, 0.0, 0.0}%
}

\NewDocumentCommand{\asya}{om}{%
    \TextFormatter[#1]{#2}{Asya}{0.0, 0.5, 0.0}%
}

\NewDocumentCommand{\gco}{om}{%
\TextFormatter[#1]{#2}{Guillaume}{0.0, 0.0, 1.0}%
}

\NewDocumentCommand{\mc}{om}{%
\TextFormatter[#1]{#2}{Matthieu}{0.5, 0, 1.0}%
}

\usepackage[accsupp]{axessibility}  %
\usepackage{xcolor,pifont}
\newcommand*\colourcheck[1]{%
  \expandafter\newcommand\csname #1check\endcsname{\textcolor{#1}{\ding{52}}}%
}
\colourcheck{blue}
\colourcheck{green}
\colourcheck{red}

\definecolor{mygray}{gray}{0.4}

\makeatletter
\@namedef{ver@everyshi.sty}{}
\makeatother
\usepackage{tikz}
\usetikzlibrary{shapes.geometric, arrows, arrows.meta}
\usepackage{pgfplots}
\pgfplotsset{width=7.5cm,compat=1.12}
\usepgfplotslibrary{fillbetween}

\begin{document}
\title{Transformer Decoders with MultiModal Regularization for Cross-Modal Food Retrieval}  %

\author{
\begin{minipage}{\linewidth}
\begin{center}
 Mustafa Shukor$^{1}$\thanks{Corresponding author: mustafa.shukor@sorbonne-universite.fr} \hspace{0.35cm} Guillaume Couairon$^{1, 2}$ \hspace{0.35cm} Asya Grechka$^{1,3}$  \hspace{0.35cm}
Matthieu Cord$^{1,4}$ \\[0.5cm] 
\scalebox{1.}{$^1$Sorbonne University\hspace{0.6cm} $^2$Meta AI \hspace{0.6cm} $^3$ Meero \hspace{0.6cm} $^4$ Valeo.ai}\\[1cm]
\end{center}
\end{minipage}
}

\maketitle
\thispagestyle{empty}

\begin{abstract}

Cross-modal image-recipe retrieval has gained significant attention in recent years. 
Most work focuses on improving cross-modal embeddings using unimodal encoders, that allow for efficient retrieval in large-scale databases, leaving aside cross-attention between modalities which is more computationally expensive. We propose a new retrieval framework, T-Food (\textbf{T}ransformer Decoders with MultiModal Regularization for Cross-Modal \textbf{Food} Retrieval) that exploits the interaction between modalities in a novel regularization scheme, while using only unimodal encoders at test time for efficient retrieval. We also capture the intra-dependencies between recipe entities with a dedicated recipe encoder, and propose new variants of triplet losses with dynamic margins that adapt to the difficulty of the task. Finally, we leverage the power of the recent Vision and Language Pretraining (VLP) models such as CLIP for the image encoder.
Our approach outperforms existing approaches by a large margin on the Recipe1M dataset. Specifically, we achieve absolute improvements of 8.1 \% (72.6 R@1) and +10.9 \% (44.6 R@1) on the 1k and 10k test sets respectively. The code is available here:\href{https://github.com/mshukor/TFood}{https://github.com/mshukor/TFood}.

\end{abstract}

\section{Introduction}
\label{sec:intro}

Multimodal learning, especially Vision and Language Pretraining (VLP), has become an attractive research field with many applications such as cross-modal retrieval \cite{plummer2015flickr30k}, Visual Question Answering (VQA) \cite{antol2015vqa} and visual reasoning \cite{xie2019visual}. An interesting playground for multimodal learning is computational cooking, which encompasses different tasks such as food categorization \cite{categ}, food perception \cite{percept}, recommendation \cite{recommendation}, or retrieval \cite{Salvador_2017_CVPR_recipe1m}.

In this work, we focus on recipe-image retrieval, which consists in retrieving the image corresponding to a given recipe and vice versa. This task has gained much attention in the recent years since the release of Recipe1M \cite{Salvador_2017_CVPR_recipe1m}, a dataset containing one million textual recipes and their corresponding images. The textual recipe is composed of 3 entities: title, ingredients, and instructions. 

Most previous works \cite{carvalho2018adamine, Salvador_2017_CVPR_recipe1m, xie2021learning_jema, wang2021cross_scan, fu2020mcen} rely on specific (dual) encoders for images and text recipes that embed the two modalities in a shared latent space. The recipe encoder usually considers the recipe entities as independent and encodes them in separate modules \cite{wang2019learning_acme}. These dual encoders are trained using pairwise, triplet or any contrastive loss optimization to align the visual and textual representations.

\begin{figure}[t]
    \centering
    \includegraphics[width=\linewidth]{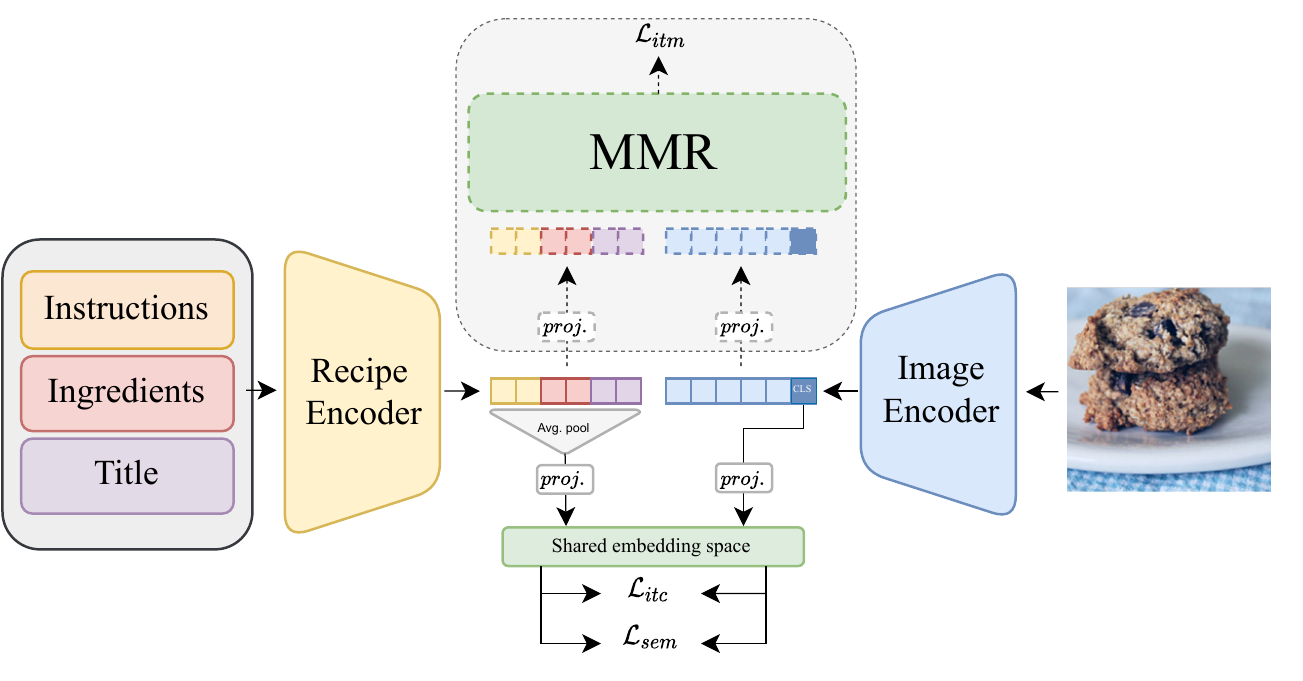}
     \caption{Illustration of our approach; the recipes and images are separately encoded by a Recipe and Image Encoder respectively. The output representation (tokens) are then projected into 2 spaces; the shared latent space where the instance and semantic losses are applied, and the multimodal space where the ITM loss is applied on top of the MMR module. The elements in dashed line are used only during training.}
    \label{fig:main}
\end{figure}

\setlength{\tabcolsep}{1pt}
\begin{table*}[ht]
    \centering
    \small
    \begin{tabular}{c|c|c|c|ccccc}
        \textbf{Title query} &
        \textbf{Ingredient query} & 
        \textbf{Instruction query} &
        \textbf{GT} &
        \multicolumn{5}{c}{\textbf{Top 5 retrieved images}} \\
        
        \hline
        \multirow{3}{*}{\tiny{Mint Chocolate Chip Frosting.}} &
        \tiny{1 cup Unsalted Butter}, & \tiny{Add sugar, cream, peppermint, and food coloring...} & \multirow{3}{*}{\includegraphics[width=0.07\linewidth, height=1.2cm]{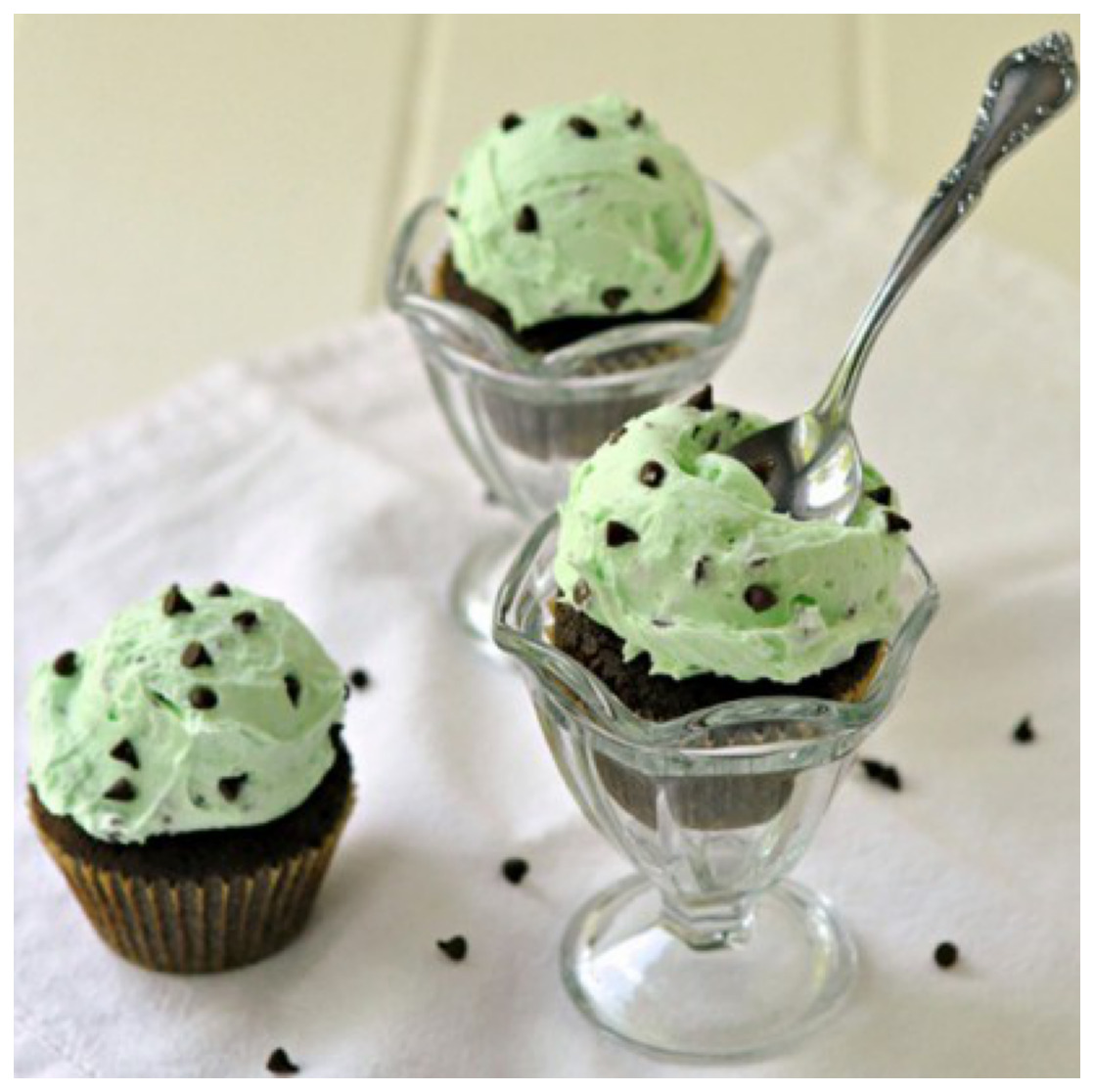}} &         \multirow{3}{*}{\includegraphics[width=0.07\linewidth, height=1.2cm]{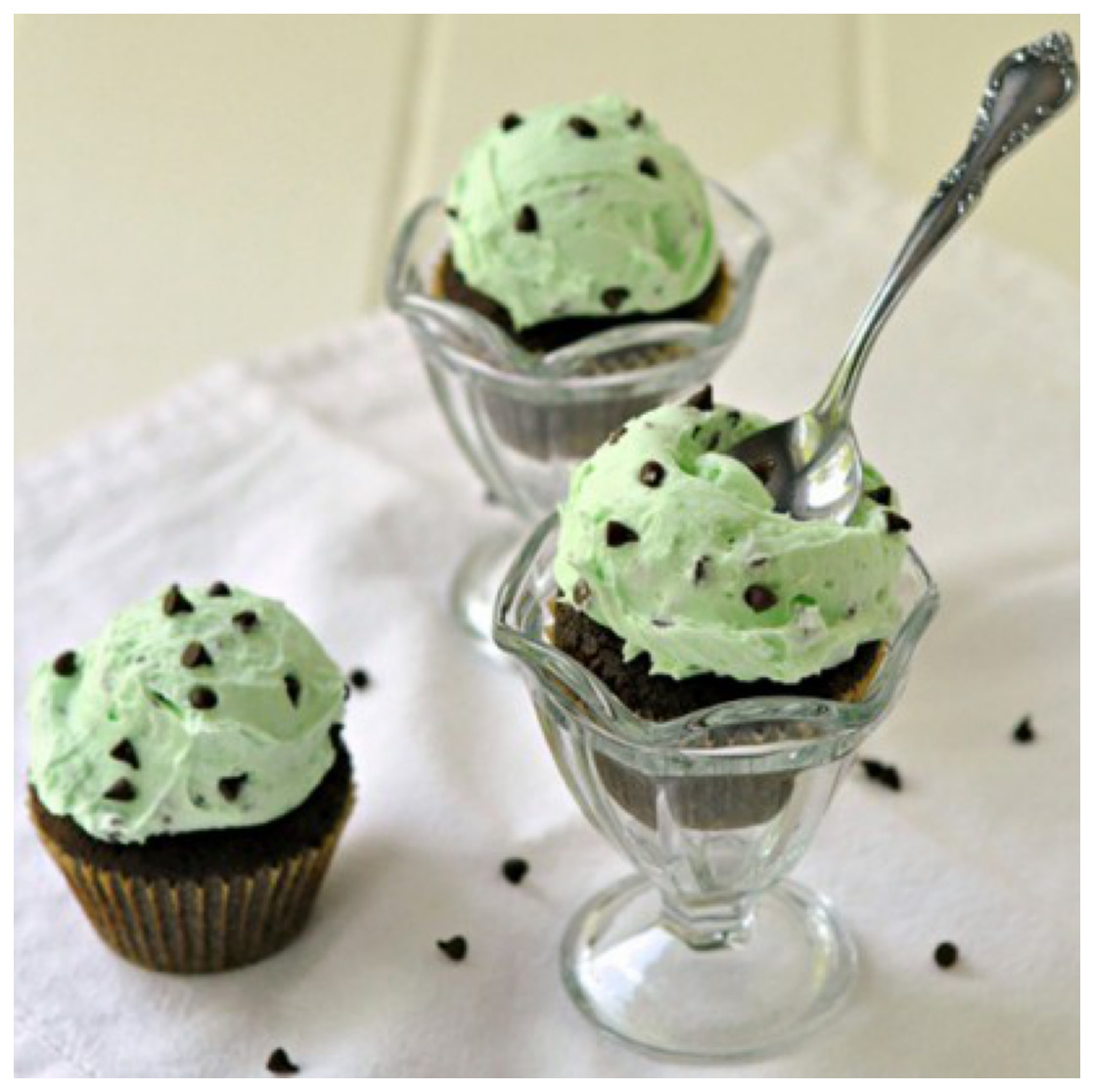}} &
        \multirow{3}{*}{\includegraphics[width=0.07\linewidth, height=1.2cm]{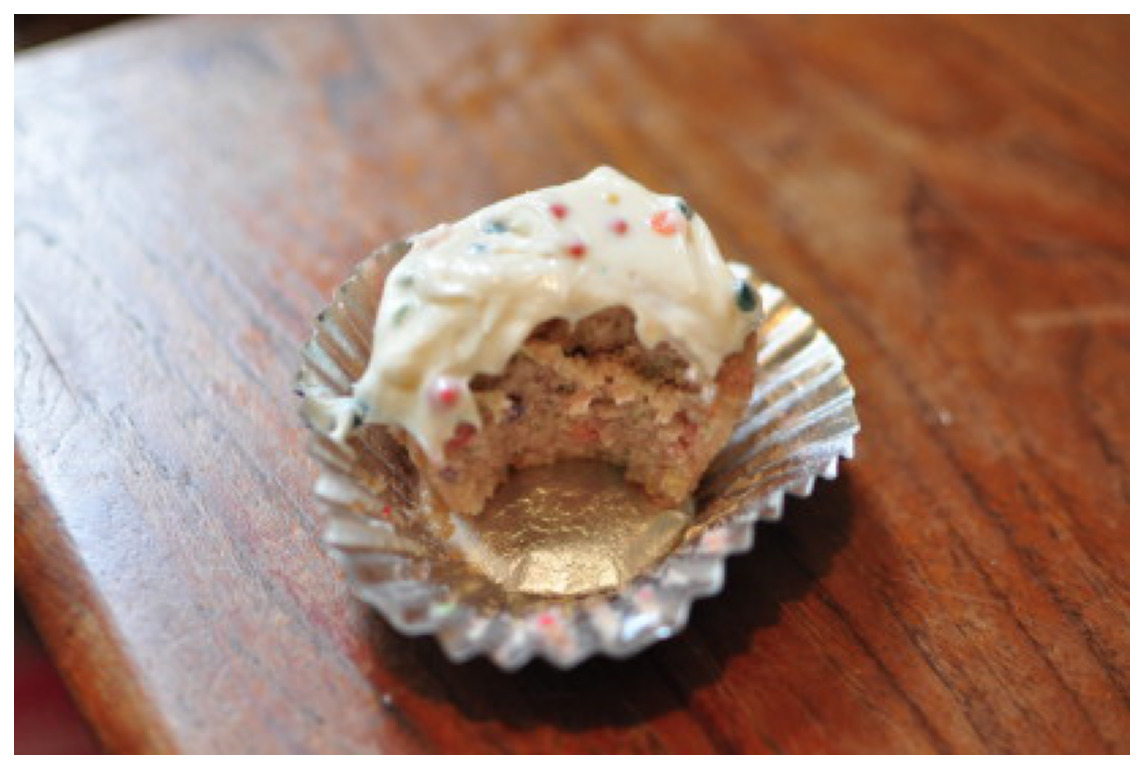}} &
        \multirow{3}{*}{\includegraphics[width=0.07\linewidth, height=1.2cm]{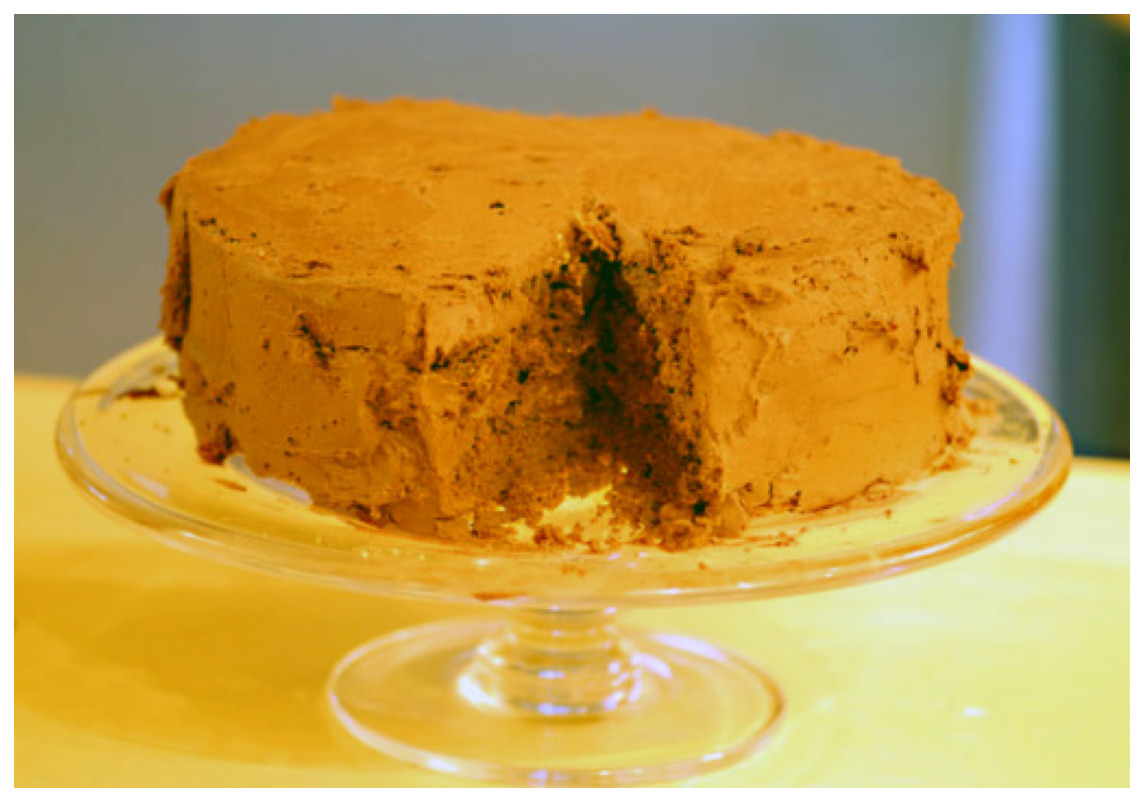}} &
        \multirow{3}{*}{\includegraphics[width=0.07\linewidth, height=1.2cm]{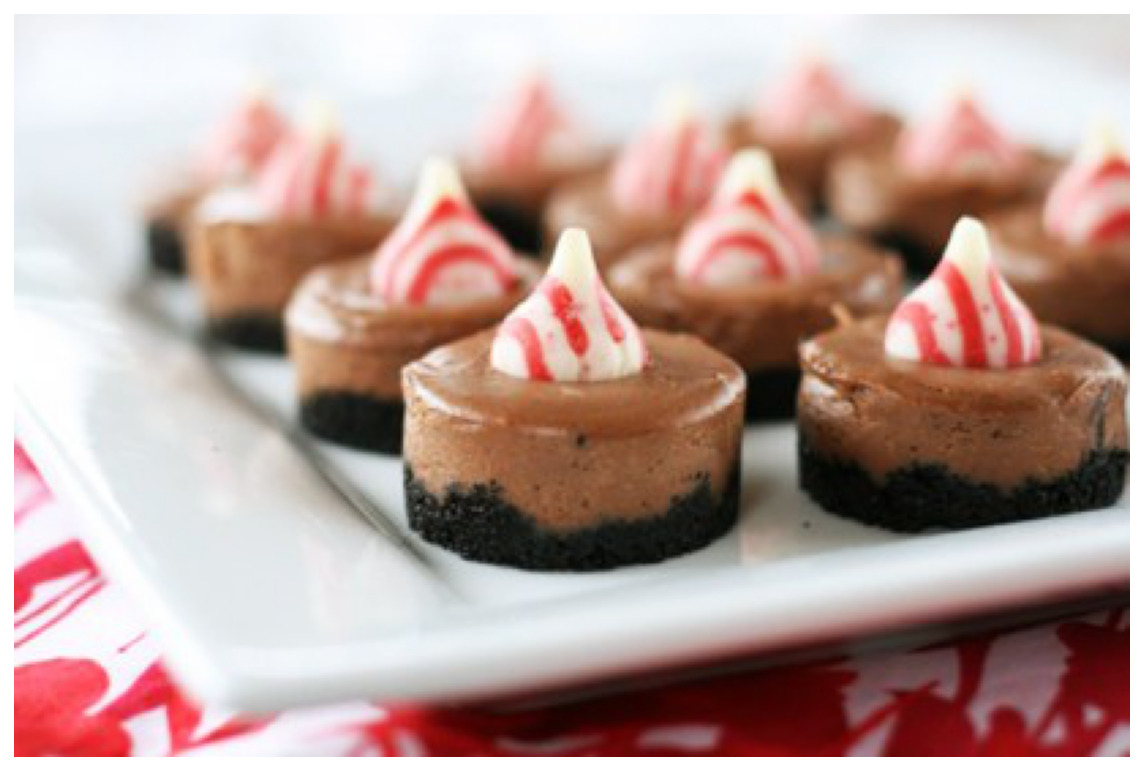}} &
        \multirow{3}{*}{\includegraphics[width=0.07\linewidth, height=1.2cm]{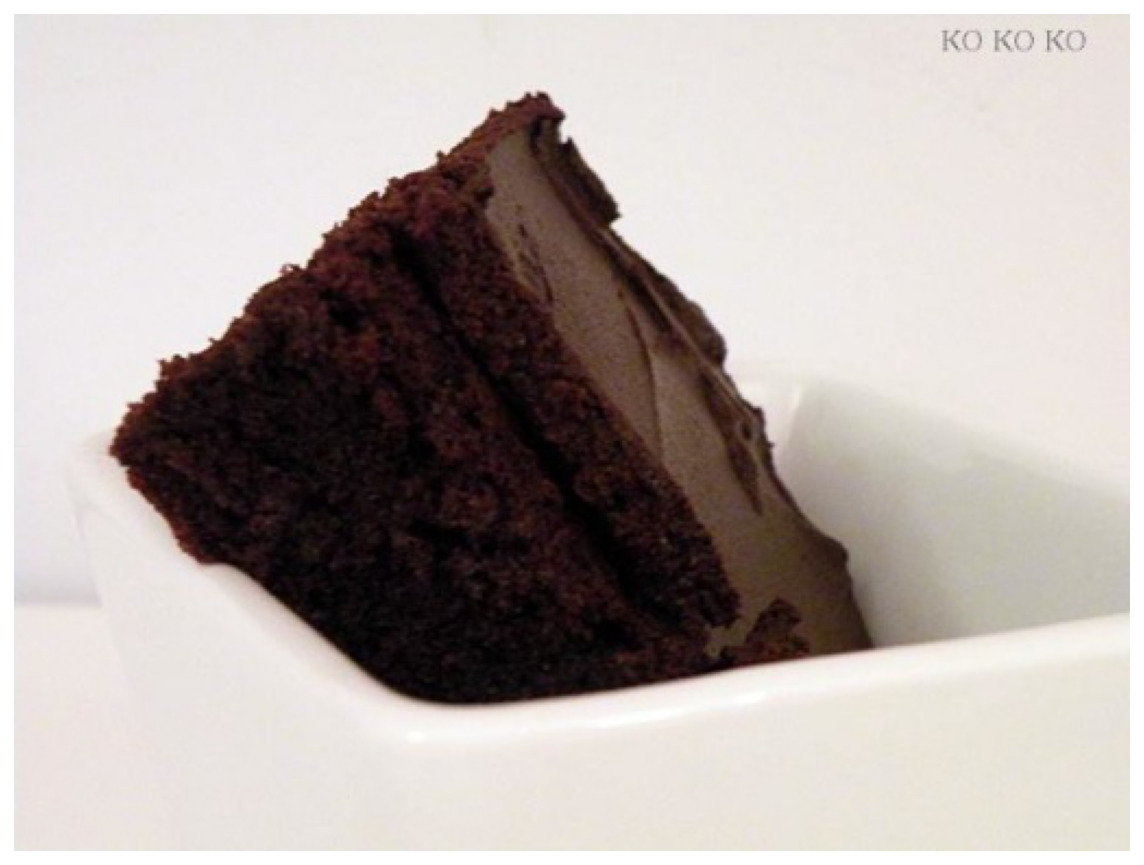}} \\
        
        & \tiny{2 Tablespoons Heavy Cream}, & \tiny{...scoop the frosting and place on top of your cupcakes} & & & & & \\
        
        & \tiny{2 drops Green Food Coloring, .. Chocolate..} & \tiny{Source: Chocolate Cupcakes with Mint Chocolate Chip ...} & & & & & \\
        \hline
        \multirow{3}{*}{\tiny{Honey-Grilled Chicken.}} &
        \tiny{1 broiler-fryer chicken, halved}, & \tiny{Place the halved chicken in a large, shallow container...} & \multirow{3}{*}{\includegraphics[width=0.07\linewidth, height=1.2cm]{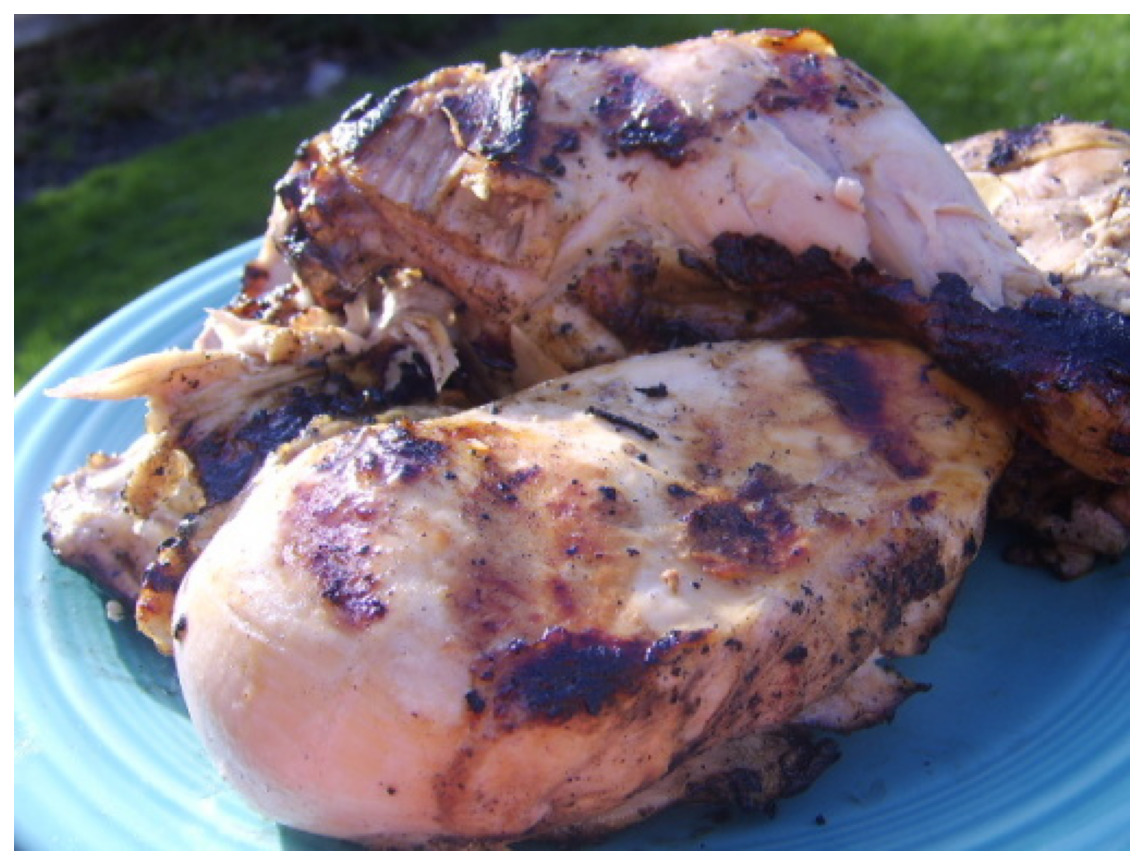}} &         \multirow{3}{*}{\includegraphics[width=0.07\linewidth, height=1.2cm]{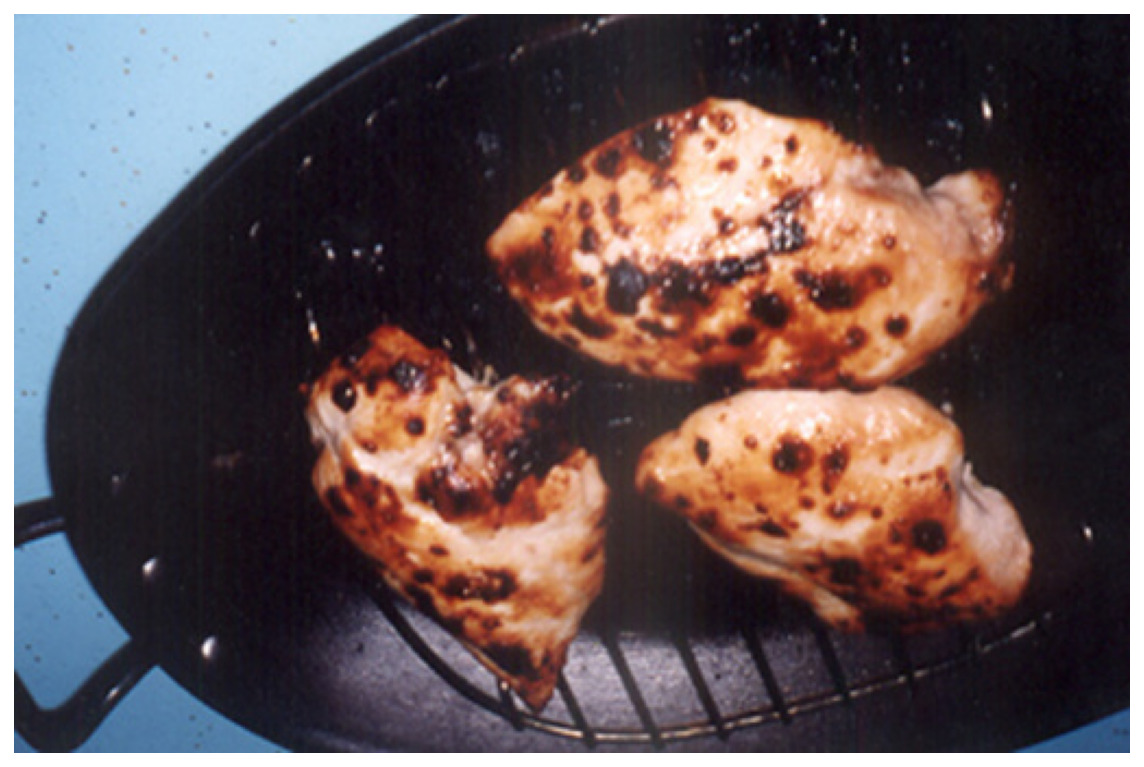}} &
        \multirow{3}{*}{\includegraphics[width=0.07\linewidth, height=1.2cm]{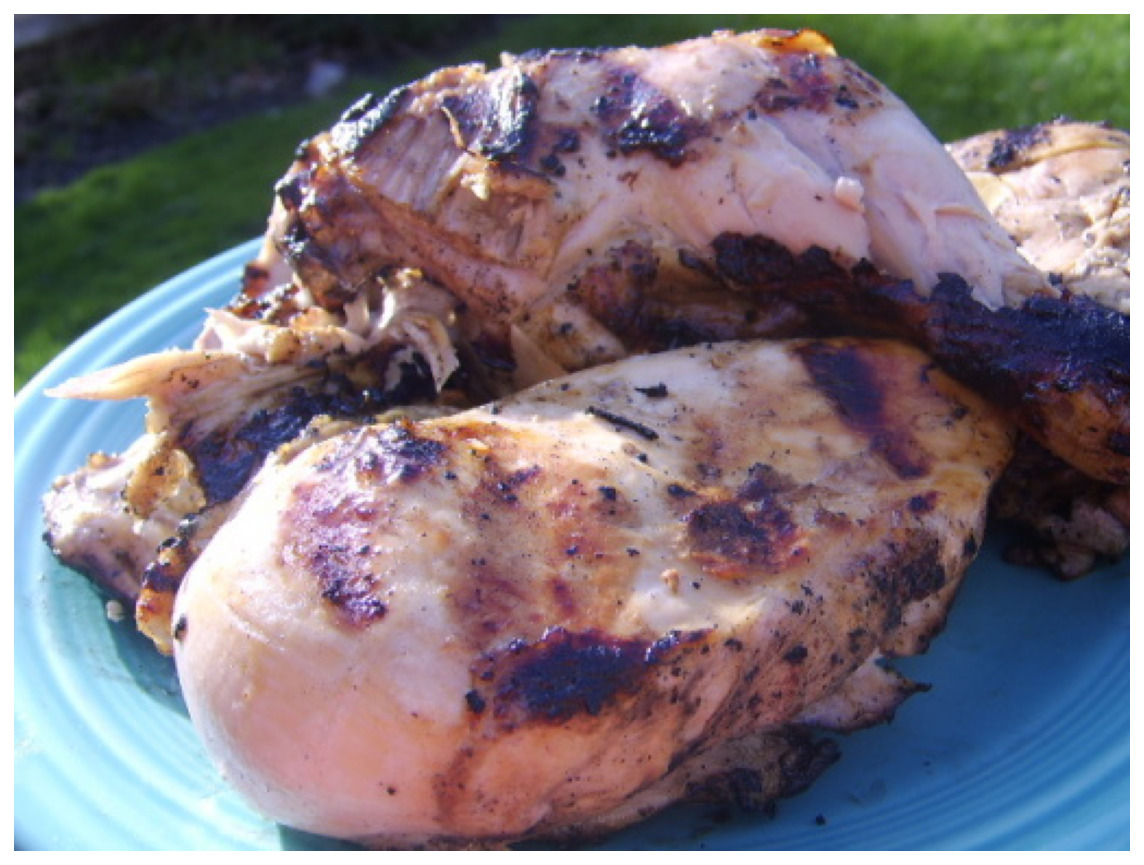}} &
        \multirow{3}{*}{\includegraphics[width=0.07\linewidth, height=1.2cm]{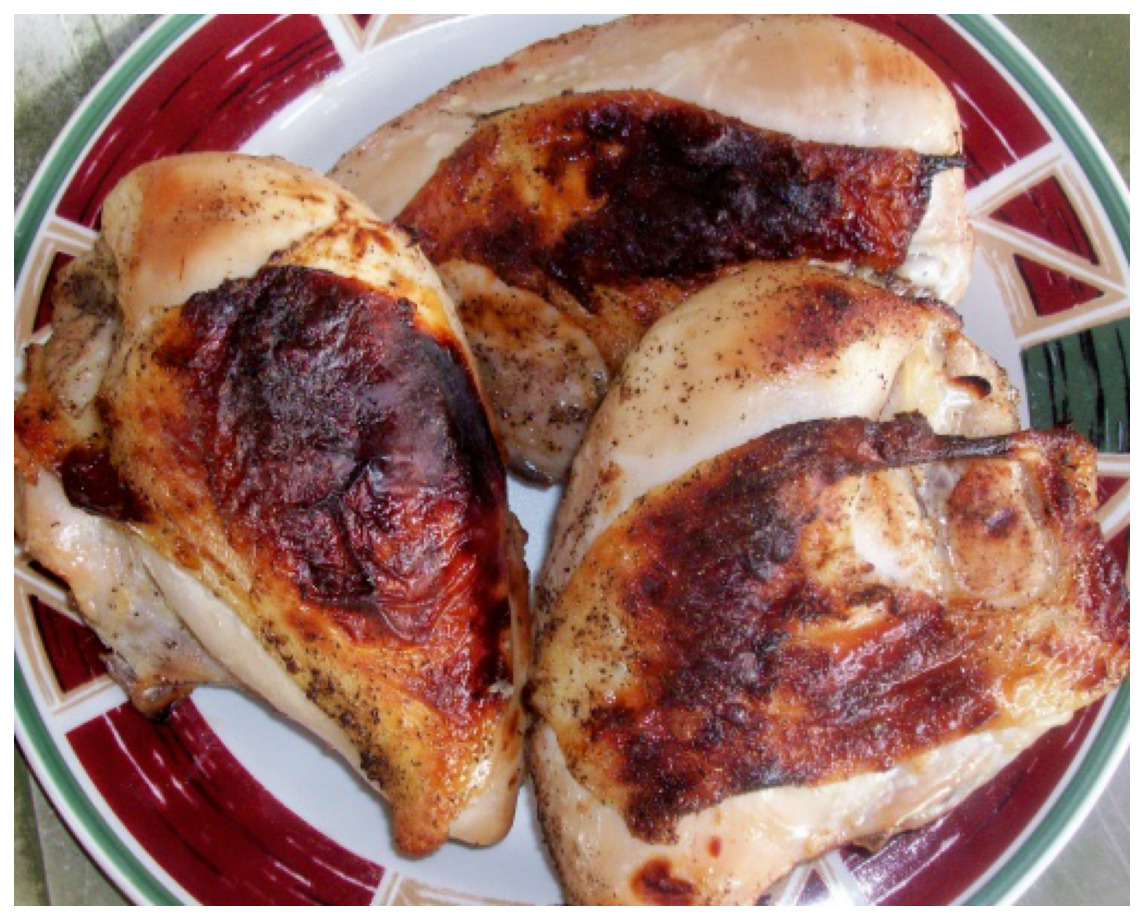}} &
        \multirow{3}{*}{\includegraphics[width=0.07\linewidth, height=1.2cm]{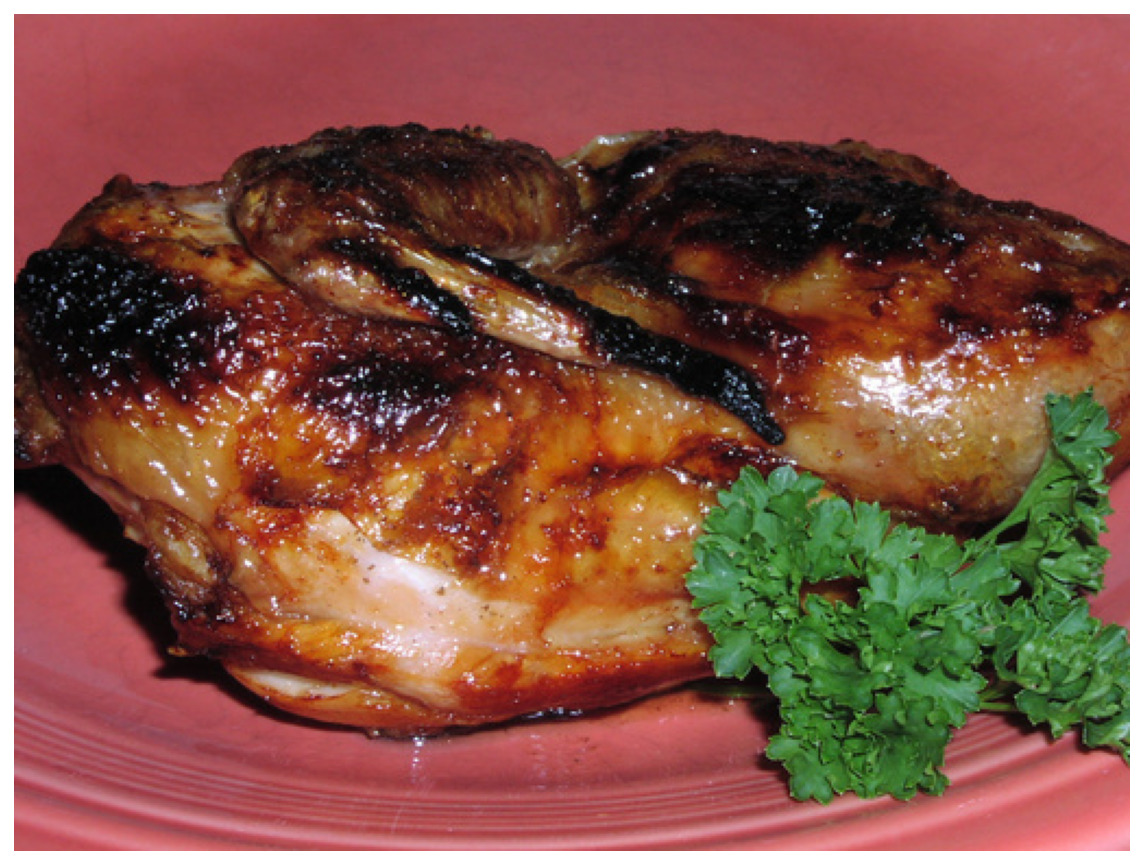}} &
        \multirow{3}{*}{\includegraphics[width=0.07\linewidth, height=1.2cm]{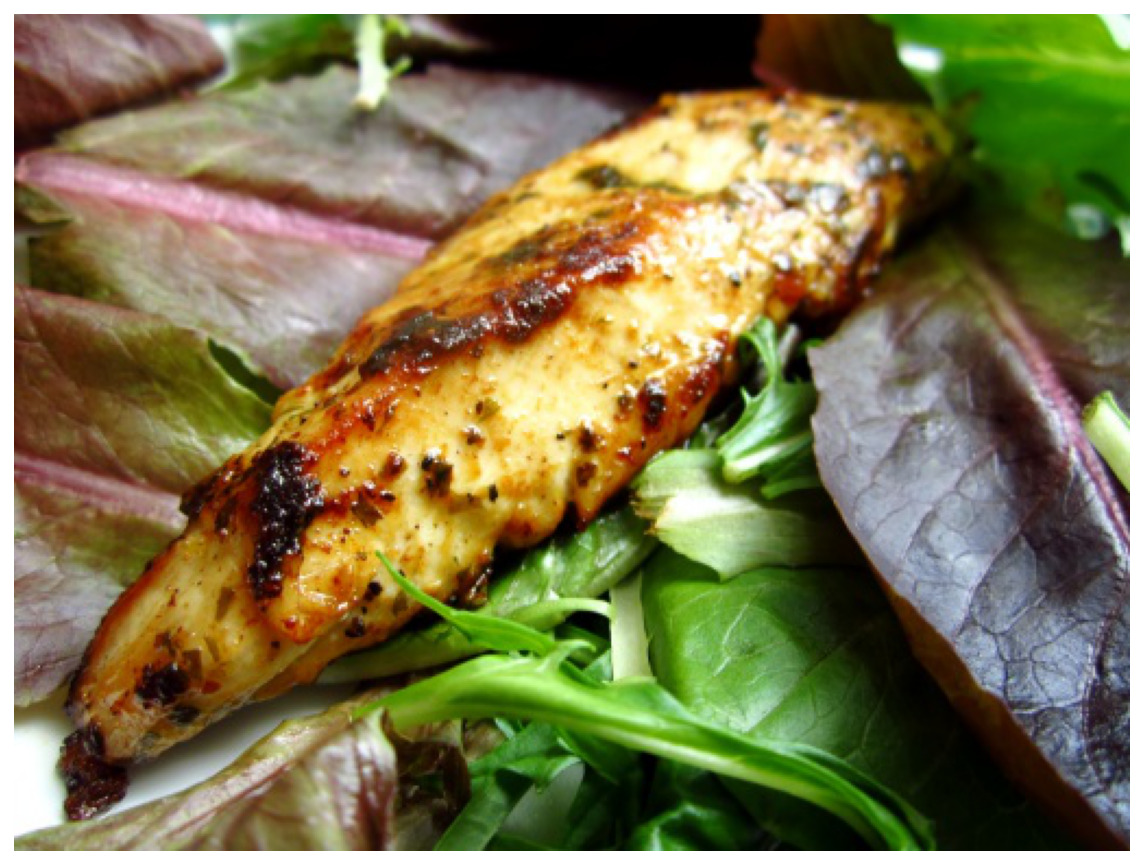}} \\
        
        & \tiny{34 cup butter, melted}, & \tiny{...Combine the remaining ingredients, stirring sauce well} & & & & & \\
        
        & \tiny{14 cup honey..} & \tiny{Grill chicken, skin side up..} & & & & & \\
\hline
        \multirow{3}{*}{\tiny{The Best Kale Ever.}} &
        \tiny{1/2 cup Kale}, & \tiny{Wash and cut kale off the stems...} & \multirow{3}{*}{\includegraphics[width=0.07\linewidth, height=1.2cm]{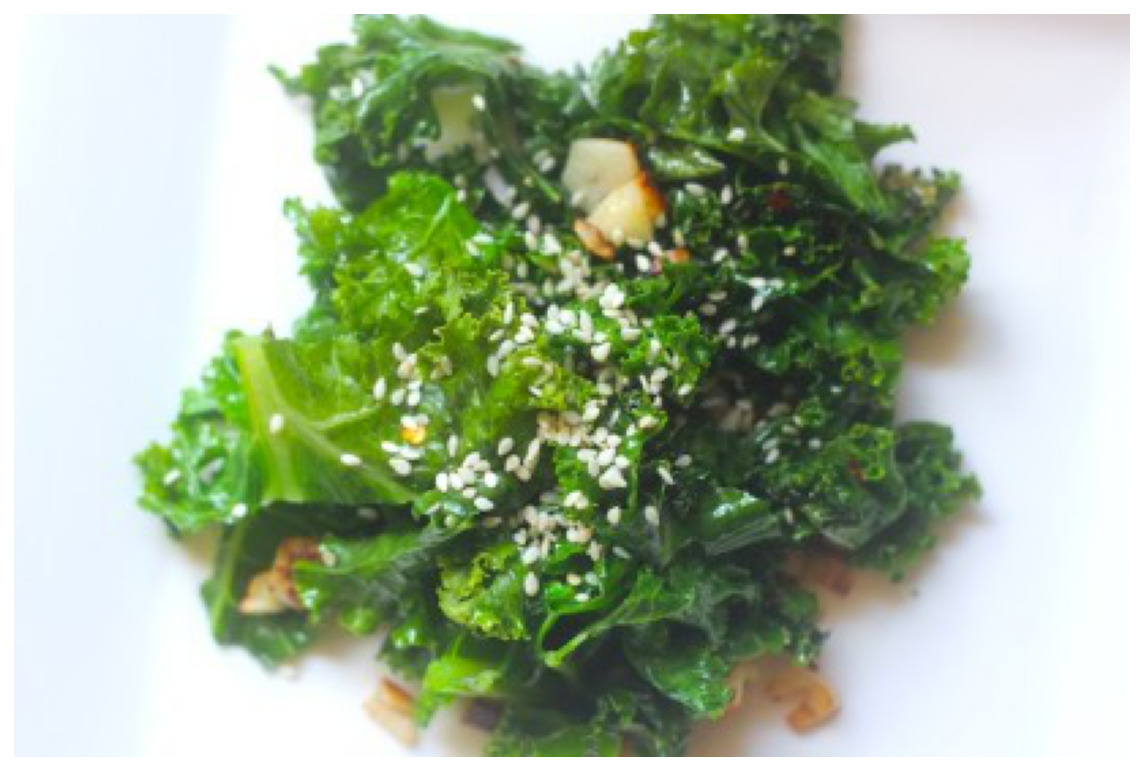}} &         \multirow{3}{*}{\includegraphics[width=0.07\linewidth, height=1.2cm]{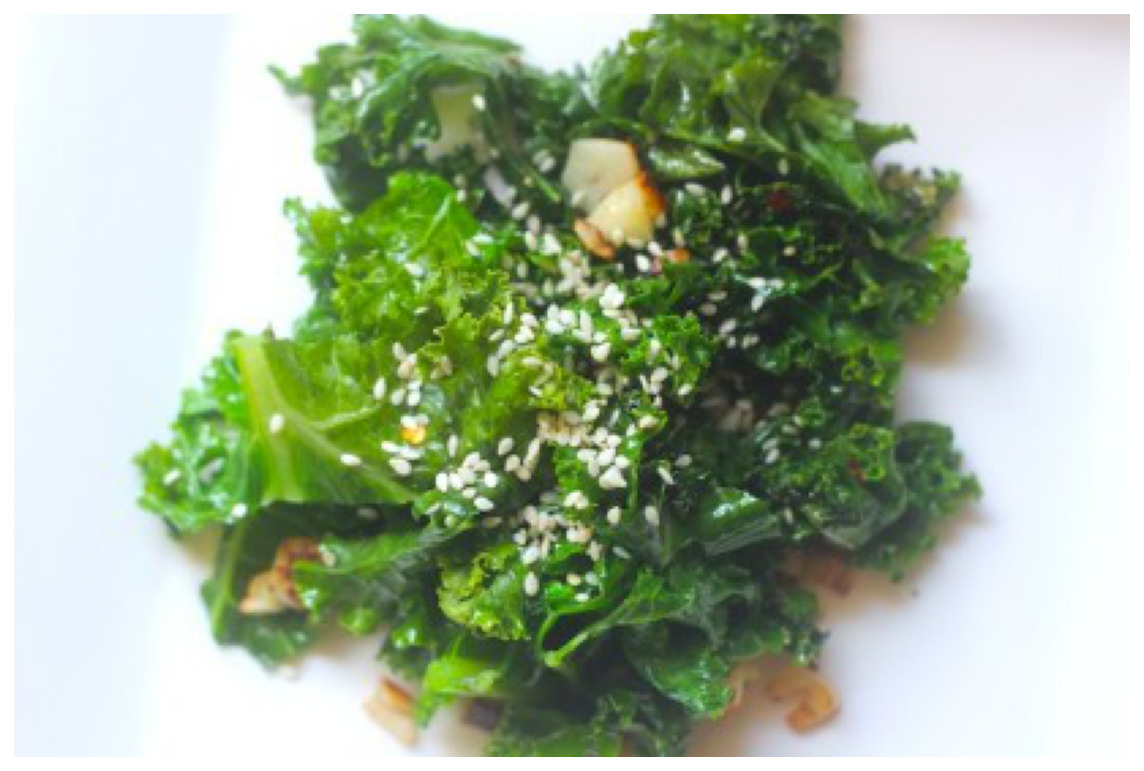}} &
        \multirow{3}{*}{\includegraphics[width=0.07\linewidth, height=1.2cm]{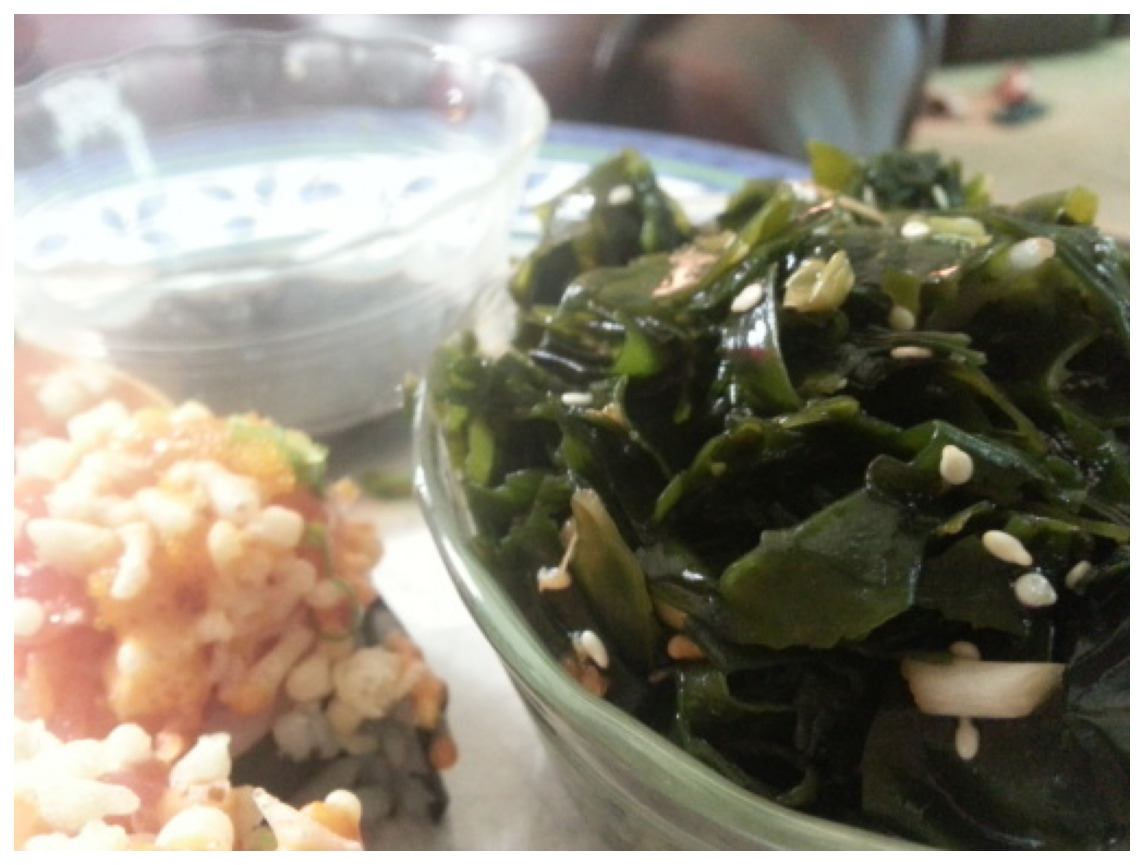}} &
        \multirow{3}{*}{\includegraphics[width=0.07\linewidth, height=1.2cm]{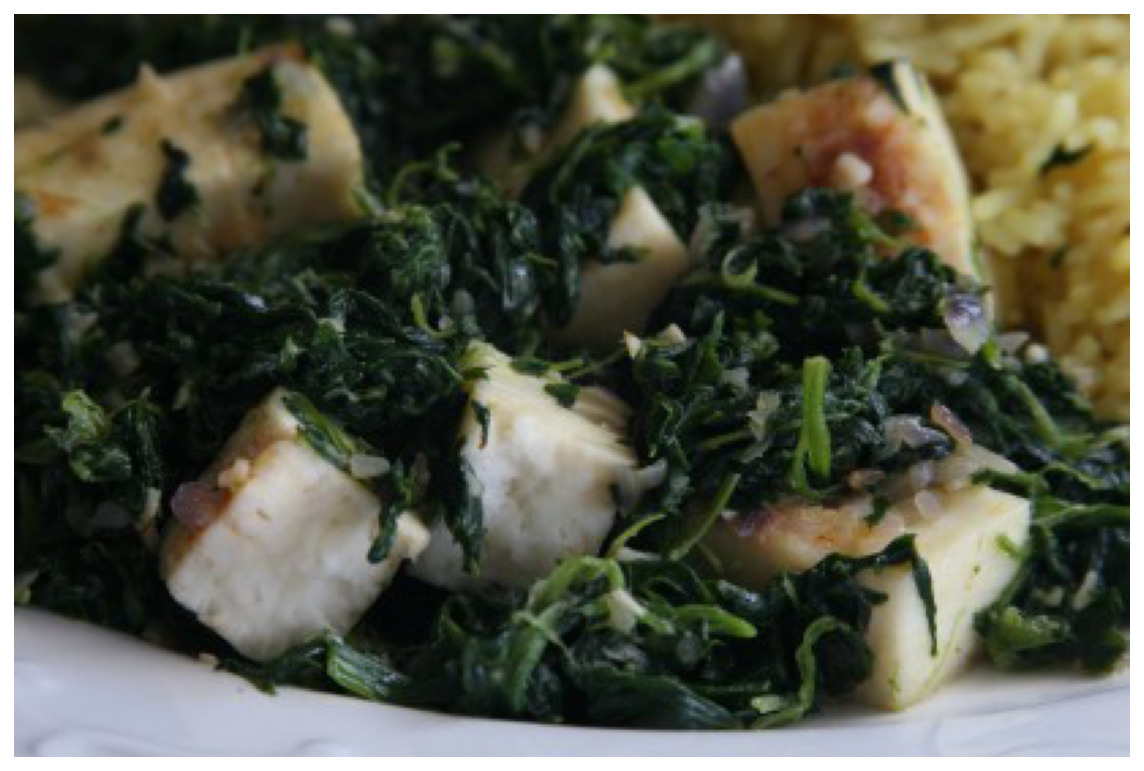}} &
        \multirow{3}{*}{\includegraphics[width=0.07\linewidth, height=1.2cm]{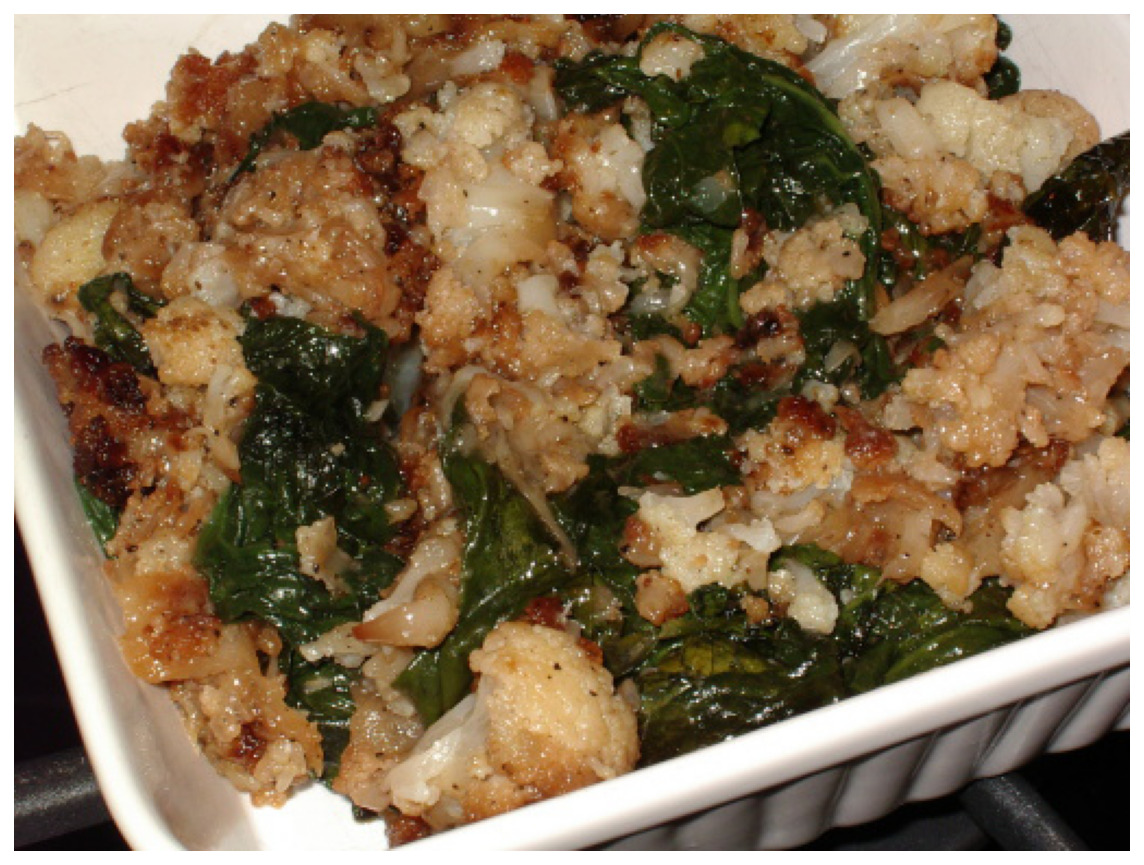}} &
        \multirow{3}{*}{\includegraphics[width=0.07\linewidth, height=1.2cm]{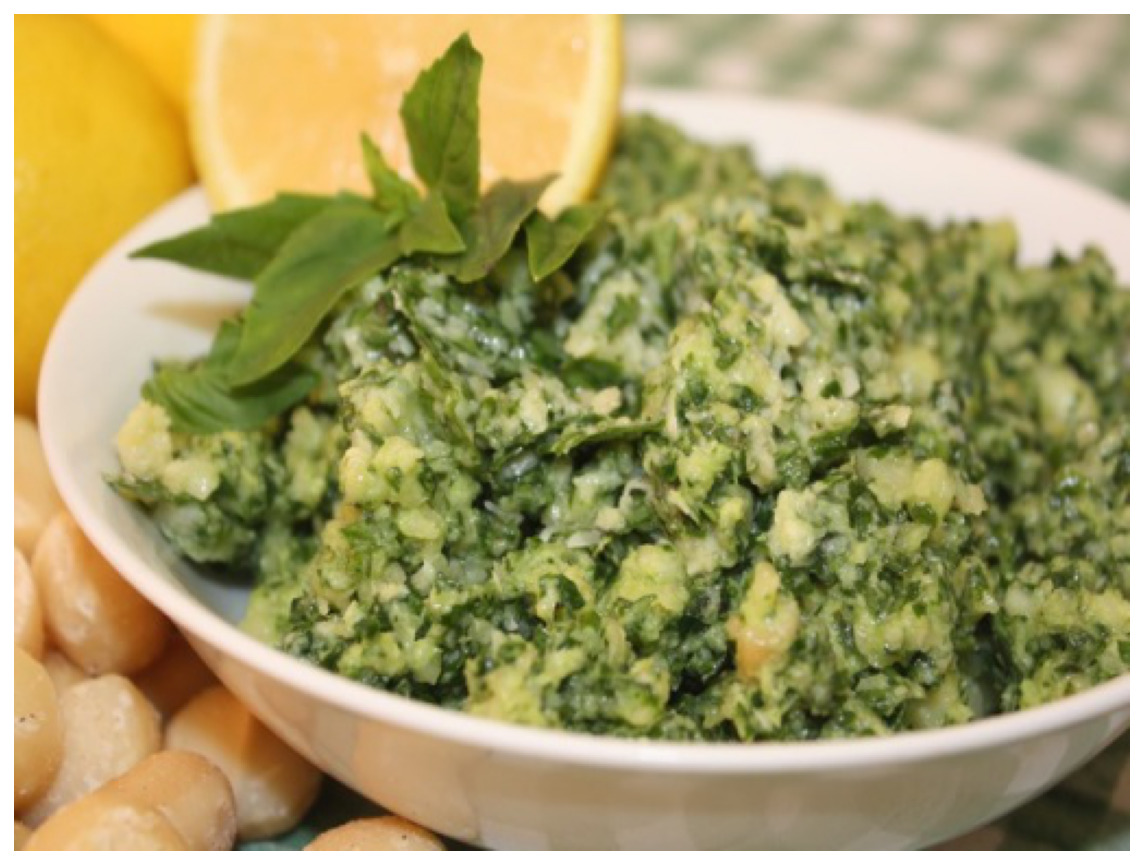}} \\
        
        & \tiny{1 teaspoon Olive Oil}, & \tiny{Heat olive oil on medium heat and add garlic} & & & & & \\
        
        & \tiny{1/4 teaspoons Red Pepper Flakes..} & \tiny{Add in kale and red pepper flakes..} & & & & & \\

        \hline

    \end{tabular}
    \caption{Qualitative results; we plot the best 5 retrieved images for each recipe query on the Recipe1M test set with 1k setup. All 5 images share semantic similarities with the query image.}
    \label{tab:visu2}
    \vspace{-0.4cm}
\end{table*}

We propose a novel strategy for cross-modal recipe retrieval based on a new architecture and learning framework. \\
Since recipe entities (title, ingredients, instructions) are highly correlated, we propose a hierarchical transformer that explicitly leverages the intra and inter entity dependencies. For the visual encoder, we consider vision transformers such as ViT, but also recent models such as CLIP-ViT pretrained on large multimodal datasets. We expect the latter to be more robust to the noise in the Recipe1M dataset which has been scraped from cooking websites. \\
For efficient large-scale retrieval, we utilize dual encoders, but we consider a more complex image-recipe interaction during the training, which we materialize by a novel transformer based module with Image-Text Matching loss that acts as a regularization to better align the encoders representation.
We also propose a new adaptive triplet loss with a dynamic margin which changes according to the difficulty of the task. The whole architecture and learning scheme is presented on Fig~\ref{fig:main}. 
Our contributions are two-fold:
\begin{itemize}
    \item Deep architecture design: (a) we propose a new recipe encoder with transformer decoders that captures the interactions between recipe entities, (b) and we leverage VLP models trained on large-scale datasets for the image encoder. (c) We complete this architecture with a multimodal block that consists of inter-connected transformers, especially designed for the training. We keep only the unimodal encoders for efficient cross-modal retrieval at test time. 
    \item Training framework: (a) We propose a novel MultiModal Regularization that consists of an Image-Text Matching loss on top of the multimodal module. (b) In addition, we introduce a new adaptive triplet loss with dynamic margin that adapts to the difficulty of the task. 
\end{itemize}

We demonstrate the interest of our approach with extensive evaluation and comparison on the Recipe1M dataset.

The work is organised as follows; we detail some of the related work in section \ref{sec:related_work}. We then present our approach in section \ref{sec:method}. The method is validated in the experiments section \ref{sec:exp}. Finally, the work is summarized in section \ref{sec:conclusion}.

\section{Related Work}
\label{sec:related_work}

\subsection{Multi Modal Learning}
Vision and Language Pretraining (VLP) has been successful for many downstream tasks such as visual question answering (VQA), image captioning, NLRVR$^2$ and cross modal retrieval \cite{chen2020uniter_uniter, huang2020pixelbert, miech2021thinking_fastslow, li2021alignalbef}. They are mainly based on dual encoders (DE) \cite{pmlr-v139-radford21a_clip, jia2021scaling_align}, multimodal encoders (ME) \cite{chen2020uniter, li2019visualbert, li2020oscar, zhou2020unified, kim2021vilt} or a combination of both \cite{lu2019vilbert, tan2019lxmert, huang2020pixel, singh2021flava, yuan2021florence}. Dual encoders, such as CLIP \cite{pmlr-v139-radford21a_clip} and ALIGN \cite{jia2021scaling_align}, use separate encoders for each modality, and a contrastive loss to align the modalities. These models are efficient during inference, however, they are trained on massive amounts of data and rely on global similarity without exploiting the fine-grained modalities interaction.

Multimodal encoders, on the other hand, employ a heavy transformer that takes the tokens of both modalities and rely on several pretraining tasks such as Masked Language Modeling (MLM) \cite{devlin2018bert}, Image-Text Matching (ITM) \cite{lu2019vilbert, chen2020uniter_uniter}, Masked Image Modeling \cite{bao2021beit, singh2021flava} or prefixLM \cite{desai2021virtex, wang2021simvlm}. Despite their success in vision-language tasks, they suffer from computational complexity due to self and cross attention between modalities, limiting their adaptability for large scale tasks. To make these models more efficient during inference, some work adopts re-ranking of top k examples after a first ranking using the dual encoders \cite{miech2021thinking_fastslow, li2021alignalbef}.
Despite the success of these approaches for cross-modal retrieval, the efficiency of the model at large scale is still a bottleneck.

\subsection{Contrastive Learning}
Contrastive learning is a wide area in representation learning, especially self supervised learning (SSL) \cite{chen2020simple_simclr, he2020momentum, chen2021exploring_simsiam, caron2020unsupervised}, where the model is forced to be invariant to input transformations. This is done by encoding the positive pairs (\emph{e.g.} the image and its transformed version) close in the latent space and the negative pairs (\emph{e.g.} different images) far away, however, while contrastive learning is becoming mostly associated with SSL, the idea goes back to distance metric learning \cite{distance_metric_learning}. 
Interestingly, the contrastive loss is not applied to the underlying representation, but after projecting it into another space. This is similar to what we propose, although, we differ in; (a) we have 2 projection modules, one linear and another more complex, (b) we work in a multimodal setting. While most of the SSL work use the InfoNCE loss \cite{oord2018representation}, here we follow other work on cross modal retrieval and use the triplet loss. Triplet loss \cite{NIPS2005_a7f592ce_triplet, ding2015deep, schroff2015facenet} takes an anchor, positive and negative example, with an objective of making the difference between the positive and negative distances are larger than a predefined margin $\alpha$. Several improvements have been proposed to select the best triplets, such as cross batch mining \cite{wang2020cross}, semi hard negatives \cite{schroff2015facenet}, batch hard triplets \cite{hermans2017defense} or truncated triplet \cite{wang2021solving}. However, little work have investigated the choice of the margin; in \cite{zhang2019learning_incremental} the authors propose multi stage training where in each stage the margin is increased. In the fashion domain, \cite{zhao2019weakly} propose adaptive margin based on the semantic distance between product descriptions.

\subsection{Cross Modal Recipe/Image Retrieval}

The task consists of retrieving the recipe corresponding to a dish image and vice versa, and it is based on computing the similarity between image and recipe embeddings in a shared latent space. Many datasets have been proposed for food computing \cite{food101, pfid, repfood, foodcam, cookpad, Salvador_2017_CVPR_recipe1m, marin2019recipe1m+}, Recipe1M and Recipe1M+ \cite{Salvador_2017_CVPR_recipe1m, marin2019recipe1m+} are the largest image/recipe datasets that consist of 800 k/13 M images respectively and 1 M recipes, structured in 3 entities; title, ingredients and instructions. It contains also the class of the recipe (\emph{e.g.}, pizza, green beans...). Despite the large number of images in Recipe1M+, Recipe1M is still only used for comparison.

Most Cross modal Image/Recipe retrieval approaches encode the recipe and the image using separate encoders \cite{Salvador_2017_CVPR_recipe1m, marin2019recipe1m+, carvalho2018adamine, dac, fu2020mcen, wang2021cross_scan, wang2019learning_acme}. The first methods rely on recurrent networks (\emph{e.g.}, LSTMs \cite{Salvador_2017_CVPR_recipe1m, carvalho2018adamine}) to encode the ingredients and instructions, which are embedded using classical approaches such word2vec \cite{w2vec} and skip-thoughts \cite{skip_thoughts}. On the image side, many work use classical image encoders such as ResNet-50 \cite{Salvador_2017_CVPR_recipe1m, carvalho2018adamine, dac, wang2019learning_acme, Zhu_2019_CVPR_r2gan}. 

Due to this challenging task, some work add discriminative or generative regularizations. Discriminative regularizations try to leverage the category or the class of the recipe in the shared latent space to obtain more semantic embeddings \cite{marin2019recipe1m+, carvalho2018adamine, wang2021cross_scan, gan_disentangling}. Generative regularizations include adversarial losses to align the distributions of the 2 modalities \cite{wang2019learning_acme, sentencebased, xie2021learning_jema} or use Generative Adversarial Networks (GANs) to reconstruct the images from either the recipe or image embedding \cite{Zhu_2019_CVPR_r2gan, gan_disentangling}, Though adversarial based approaches show good results, the discriminative ones remain much simpler and more efficient to train.

Ingredients and instructions do not contribute equally in the alignement, some ingredients can be seen clearly in the image while others does not appear. This motivates some work to include attention modules in the recipe encoder \cite{wang2021cross_scan, fu2020mcen, hybrid_fusion} or the image encoder \cite{fu2020mcen} to weight differently several tokens/regions when fusing the 2 modalities. Seeing the success of image transformers in vision \cite{vit, touvron2021training, liu2021swin} and text \cite{devlin2018bert}, some work started recently to use transformers, showing promising results \cite{salvador2021revamping, guerrero2021cross_xmrs}. 

Despite being successful for image and language tasks, multimodal transformers are still little investigated in the cooking context. Few work have tried to  use cross attention modules between the 2 modalities \cite{multisubspace, hybrid_fusion}, or a mulimodal transformer encoder \cite{trans_encod_hybrid}, but their performance lag behind the SOTA work.

\section{T-Food framework}
\label{sec:method}
Given a dataset composed of image and recipe pairs, we propose to align the representations of the two modalities for cross-modal retrieval. Fig. \ref{fig:main} illustrates the approach.
First, each modality is encoded separately into two sequences of token embeddings. Notably, the recipe encoder fuses the information of \textit{title, ingredients} and \textit{instructions} into a single sequence.
These two sequences are used in two different manners: (1) we extract two fixed-size embeddings that are used for retrieval at test time, and are brought closer together with a triplet loss at train time. (2) we feed the two sequences into the MultiModal Regularization (MMR) module, which computes a fine-grained alignement score (Image-Text Matching loss) between the input image and recipe.

\subsection{Dual Encoders (DE)}
In this section, we detail the image and recipe encoders (\emph{i.e.}, Dual Encoders (DE)).

\begin{figure}[t]
    \centering
    \includegraphics[width=\linewidth]{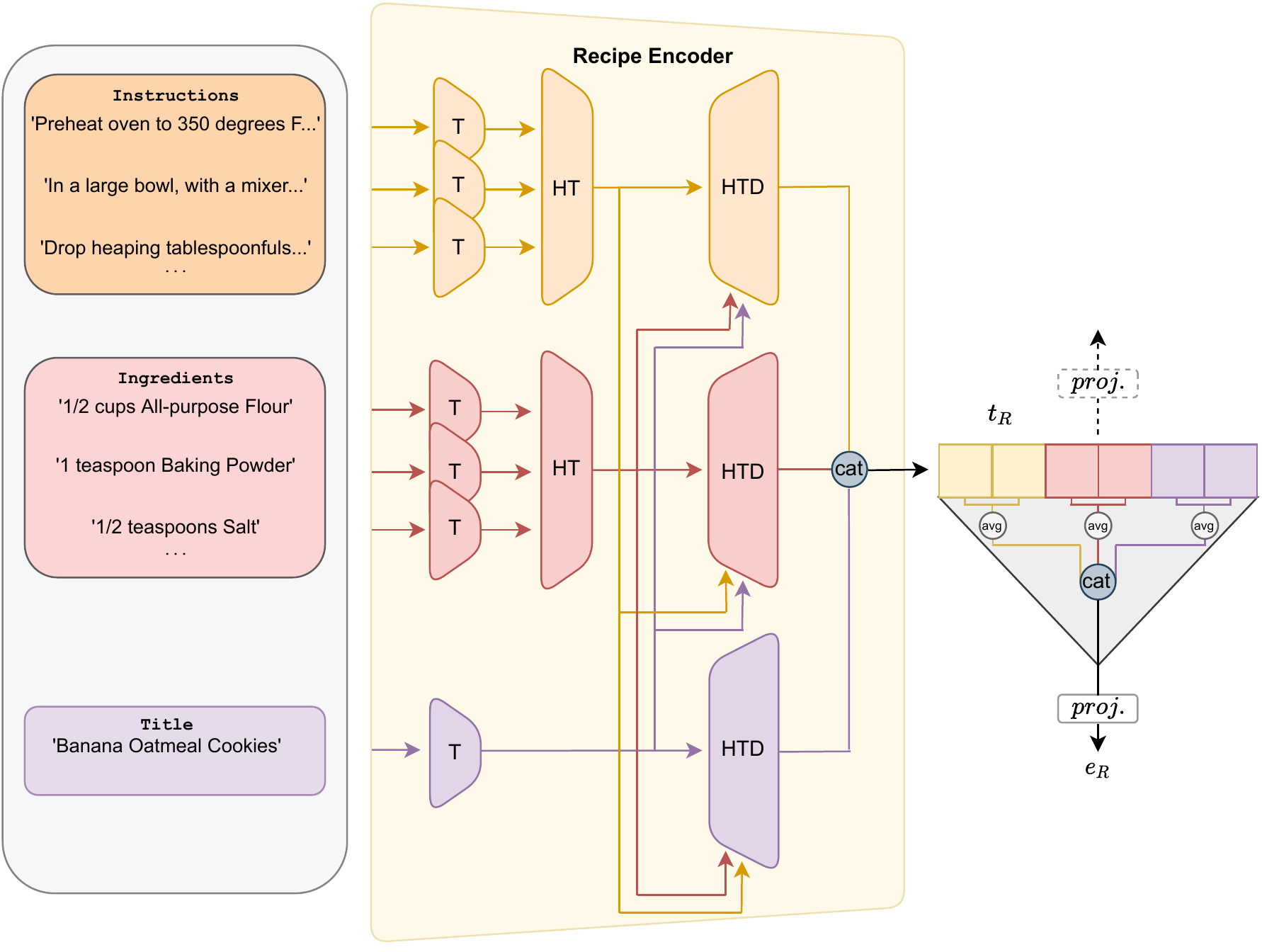}
    \caption{Illustration of the Recipe Encoder: the list of ingredients, instructions and the title are fed into transformer encoders $T$, then the list of output tokens are fed into other transformer encoders $HT$ (except for the title). The tokens of each entity are then processed as Query by transformer decoders $HTD$ that takes the other 2 entities as Keys and Values. The output tokens $t_R$ are concatenated and linearly projected, and the recipe embedding $e_R$ is obtained by averaging then concatenating and linearly projecting the output tokens of each recipe entity.}
    \label{fig:re}
\end{figure}
\paragraph{Image Encoder}
\label{sec:image_encoder}
Inspired by the success of vision transformers \cite{vit, touvron2021training, liu2021swin} and particularly for the underlying task of cross-modal retrieval \cite{salvador2021revamping}, we propose to use a vision transformer (\emph{i.e.}, ViT B/16) as an image encoder.
In addition, fine-tuning large vision/language models on downstream tasks has been shown to be an effective strategy \cite{chen2020uniter_uniter,shen2021much}, thus we propose to fine-tune the CLIP ViT B/16 \cite{pmlr-v139-radford21a_clip} which was trained on a large-scale dataset scraped from the internet.
Specifically, the image encoder $E_I$ encodes an input image $I$ into a sequence of image tokens $t_I = E_I(I)$ along with a [CLS] token output that is linearly projected to form a fixed-size image embedding $e_I$.

\paragraph{Recipe Encoder}
Different from other works that use  recurrent networks and pretrained embeddings, we follow \cite{salvador2021revamping} using hierarchical transformers to encode the recipes as raw text. 
As illustrated in Fig. \ref{fig:re}, each tokenized sentence in the list of ingredients ($t_{ing}^1$), instructions ($t_{ins}^1$) and the title $t_{ttl}^1$ is processed by a transformer $T$. For the ingredients and instructions, another transformer $HT$ processes the output of the first one (\emph{i.e.} the sentence embeddings), in a similar way to obtain the embeddings $t_{ins}^2= HT(T(t_{ins}^1))$ and $t_{ing}^2= HT(T(t_{ing}^1))$ for instructions and ingredients respectively. For the title, the embedding is obtained from the first transformer $t_{ttl}^2 = T(t_{ttl}^1)$.

However, we argue that encoding each recipe entity separately, as done in \cite{salvador2021revamping}, is not effective to capture the interactions between them. To this end, we propose to add another level of hierarchy to capture the interactions between all element tokens that leads to better intra-fusion. Specifically, a transformer decoder with self and cross attention $HTD(Q, K, V)$ takes the output tokens of each entity as a query $Q$ and the concatenation of the tokens of the other 2 remaining entities as keys $K$ and values $V$, the final tokens are obtained by concatenating all the tokens:
\begin{align}
t_{ttl}^3 & = HTD(t_{ttl}^2, [t_{ing}^2;t_{ins}^2], [t_{ing}^2;t_{ins}^2]), \nonumber \\
t_{ing}^3 & = HTD(t_{ing}^2, [t_{ttl}^2;t_{ins}^2], [t_{ttl}^2;t_{ins}^2]), \nonumber \\
t_{ins}^3 & = HTD(t_{ins}^2, [t_{ttl}^2;t_{ing}^2], [t_{ttl}^2;t_{ing}^2]), \nonumber \\
t_R & = [t_{ttl}^3; t_{ing}^3; t_{ins}^3]. 
\end{align}
Finally the output tokens of each recipe entity are averaged, concatenated and linearly projected to obtain the final recipe embedding $e_R$.

\begin{figure}[t]
    \centering
    \includegraphics[width=\linewidth, height=0.6\linewidth]{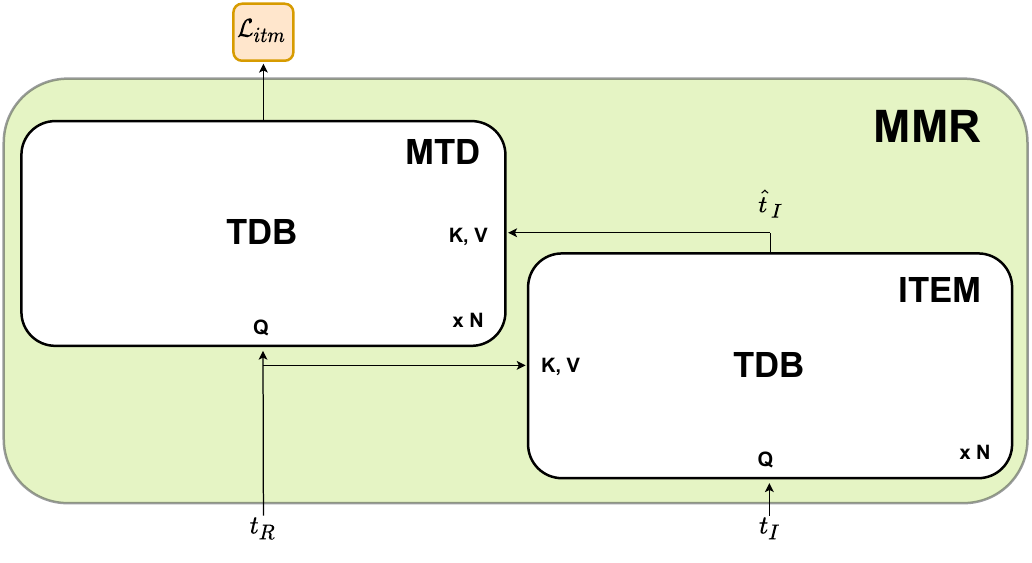}
    \caption{Illustration of our MultiModal Regularization (MMR); The Transformer Decoders Block (\textbf{TDB}) consists of self attention, cross attention, feed forward and layer normalization layers with residual connections, which are repeated N times. The cross attention takes the tokens of one modality as Queries (Q) and the tokens of the other modality as Keys (K) and Values (V). The image tokens are enhanced by the ITEM module by attending to the recipe tokens. The recipe tokens are fed to MTD as Q and the enhanced image tokens as K and V. After fusing the two modalities an ITM loss is applied. Note that this module is used only during training}
    \label{fig:MR}
\end{figure}

\subsection{MultiModal Regularization (MMR)}

Inspired by other work on vision and language pretraining \cite{chen2020uniter_uniter, li2021alignalbef, singh2021flava}, we propose a MultiModal Regularization (Fig. \ref{fig:MR}) that consists of transformer decoders \cite{vaswani2017attention} with an Image-Text Matching loss (ITM) \cite{chen2020uniter_uniter} to further guide the dual encoders during the alignment process. Different from other work that use multimodal encoders during training and testing, here our multimodal modules are used only during training. Compared to other representation learning approaches, such as self-supervised learning, \cite{chen2020simple_simclr} where a simple projection layer is added on top of the feature representation to compute the contrastive loss, here, this projection layer is replaced by a more complicated multimodal module, and the contrastive loss is replaced by ITM. 

The main block of MMR consists of a transformer decoder which particularly uses cross attention: the Queries (Q) come from one modality while the Keys (K) and Values (V) come from the other modality (we do not distinguish between K and V, which denoted as KV). In other words, the queries are updated by taking into account the dependencies between the 2 modalities. Specifically, each vector of Q is updated by taking a weighted sum of the vectors of KV of the other modality, where each weight is the similarity of each one of these KV vectors to the underlying Q vector. Our choice for using transformer decoders is based on their computation efficiency compared to that of a transformer encoder, which takes all the tokens as Q, V and K. In addition,  we choose the recipe tokens ($t_R$) as queries, as the number of recipe tokens ($N_R \sim 45$) is much less than the number of image tokens ($N_I=196$), which reduces the complexity to ($\sim N_R^2 + N_R \times N_I$) in contrast to ($\sim (N_R+N_I)^2$) for a transformer encoder.

The MMR module consists of inter-connected transformers; an Image Tokens Enhancement Module (ITEM) to enhance the image tokens, followed by a multimodal transformer decoder (MTD):
\paragraph{Image Tokens Enhancement Module (ITEM)}
We propose to enhance the image tokens before entering MTD, by adding a transformer decoder that takes the image tokens as queries (Q) and the recipe tokens as Keys and Values (KV):
\begin{equation}
    \hat{t}_I = ITEM(t_R, t_I).
\end{equation}
This module enriches the image tokens by attending to the recipe elements. While it seems natural to use only the ingredients, as they are present in different regions/tokens of the image, we found that other recipe elements, even if they are not explicitly presented in the image, brings additional improvements (see supplementary material).

\setlength{\tabcolsep}{7pt}
\begin{table*}[th]
\centering
\resizebox{\textwidth}{!}{
\begin{tabular}{@{}l|cccc|cccc|cccc|cccc@{}}
\toprule
\multirow{3}{*}{} & \multicolumn{8}{c|}{\textbf{1k}}                                                    & \multicolumn{8}{c}{\textbf{10k}} \\ \cmidrule(l){2-17} 
                  & \multicolumn{4}{c|}{\textbf{image-to-recipe}} & \multicolumn{4}{c|}{\textbf{recipe-to-image}} & \multicolumn{4}{c|}{\textbf{image-to-recipe}} & \multicolumn{4}{c}{\textbf{recipe-to-image}} \\ \cmidrule(l){2-17} 
                  & medR     & R@1      & R@5    & R@10    & medR     & R@1     & R@5     & R@10    & medR     & R@1      & R@5    & R@10    & medR     & R@1     & R@5     & R@10    \\ \midrule
Salvador et al. \cite{Salvador_2017_CVPR_recipe1m}    & 5.2      & 24.0    & 51.0      &   65.0     &   5.1       & 25.0       &     52.0   &       65.0 & 41.9    & -     &  -     &   -     &      39.2    &      -  &     -   &    -    \\
Adamine \cite{carvalho2018adamine}   &    2.0      &   40.2     &   68.1     &   78.7  &    2.0      & 39.8    &  69.0     &    77.4     &      13.2    &   14.8     &    34.6    &    46.1 & 14.2     & 14.9    &   35.3    &  45.2   \\
R2GAN \cite{Zhu_2019_CVPR_r2gan}                  &     2.0     &     39.1    &   71.0    &  81.7      &  2.0        &    40.6    &    72.6    &       83.3 &  13.9        &   13.5      &   33.5    &     44.9   &     12.6     &   14.2     &   35.0     &   46.8     \\
MCEN \cite{fu2020mcen}               &   2.0       &   48.2      &    75.8   &      83.6  &    1.9      &   48.4     &  76.1      &    83.7    &  7.2        &      20.3   &    43.3   &   54.4     &     6.6     &   21.4     &  44.3      &    55.2    \\
ACME \cite{wang2019learning_acme}              &    1.0      &   51.8      &   80.2    &    87.5    &   1.0       &    52.8    &   80.2     &       87.6 &     6.7     &   22.9     &  46.8     &    57.9    &      6.0    &  24.4      &    47.9    &      59.0  \\
SN \cite{sentencebased}               &    1.0      & 52.7        &     81.7  &    88.9    &    1.0      &  54.1      &  81.8      &   88.9     &    7.0      &    22.1    &    45.9   &    56.9    &    7.0      &     23.4   &  47.3      &  57.9      \\
IMHF \cite{intra_inter}               &    1.0      & 53.2        &     80.7  &    87.6    &    1.0      &  54.1      &  82.4      &   88.2     &    6.2      &    23.4    &    48.2   &    58.4    &    5.8      &     24.9   &  48.3      &  59..4      \\
\textit{Wang et. al}  \cite{wang2021learning}               &    1.0      & 53.5        &     81.5  &    88.8    &    1.0      &  55.0      &  82.0      &   88.8     &    6.0      &    23.4    &    48.8   &    60.1    &    5.6      &     24.6   &  50.0      &  61.0      \\
SCAN \cite{wang2021cross_scan}               &    1.0      & 54.0        &     81.7  &    88.8    &    1.0      &  54.9      &  81.9      &   89.0     &    5.9      &    23.7    &    49.3   &    60.6    &    5.1      &     25.3   &  50.6      &  61.6      \\
HF-ICMA \cite{hybrid_fusion}               &    1.0      & 55.1        &     86.7  &    92.4    &    1.0      &  56.8      &  87.5      &   93.0     &    5.0      &    24.0    &    51.6   &    65.4    &    4.2      &     25.6   &  54.8      &  67.3      \\
MSJE \cite{xie2021learning}               &    1.0      & 56.5        &     84.7  &    90.9    &    1.0      &  56.2      &  84.9      &   91.1     &    5.0      &    25.6    &    52.1   &    63.8    &    5.0      &     26.2   &  52.5      &  64.1      \\

SEJE \cite{xie2021learning2}               &    1.0      & 58.1        &     85.8  &    92.2    &    1.0      &  58.5      &  86.2      &   92.3     &    4.2      &    26.9    &    54.0   &    65.6    &    4.0      &     27.2   &  54.4      &  66.1      \\

M-SIA \cite{multisubspace} &  1.0       &     59.3    &    86.3   &    92.6    &     1.0     &     59.8   &    86.7    &      92.8  &     4.0    &    29.2     &    55.0   &    66.2    &    4.0      &     30.3   &     55.6   &  66.5     \\

DaC \cite{dac}             &  1.0       &     60.2    &    84.0   &    89.7    &     -     &     -   &    -    &      -  &     4.0    &    30.0     &    56.5   &    67.0    &    -      &     -   &     -   &        \\

X-MRS \cite{guerrero2021cross_xmrs}             &  1.0       &     64.0    &    88.3   &    92.6    &     1.0     &     63.9   &    87.6    &      92.6  &     3.0    &    32.9     &    60.6   &    71.2    &    3.0      &     33.0   &     60.4   &  70.7       \\

H-T \cite{salvador2021revamping}             &  1.0       &     60.0    &   87.6   &     \underline{92.9}    &     1.0     &     60.3   &    87.6    &      93.2  &     4.0    &    27.9     &    56.4   &    68.1    &    4.0      &     28.3   &     56.5   &  68.1       \\\midrule
H-T (ViT)           &  1.0       &     64.2    &    \underline{89.1}   &    \textbf{93.4}    &     1.0     &     64.5   &    \underline{89.3}    &      \textbf{93.8}  &     3.0    &    33.5     &    62.1   &    72.8    &    3.0      &     33.7   &     62.2   &  72.7       \\
 \midrule
T-Food (ViT)           & \textbf{1.0} & \underline{68.2}  & 87.9  &91.3  & \textbf{1.0} & \underline{68.3}  &87.8  & 91.5 & \textbf{2.0}      & \underline{40.0}          & \underline{67.0}  &\underline{75.9}    &\textbf{2.0}   & \underline{41.0}       &\underline{67.3}         &      \underline{75.9}            \\ 
\textcolor{mygray}{T-Food (CLIP-ViT)}            &  \textcolor{mygray}{\textbf{1.0}}   &  \textcolor{mygray}{\textbf{72.3}}  & \textcolor{mygray}{\textbf{90.7}}    & \textcolor{mygray}{\textbf{93.4}}  & \textcolor{mygray}{\textbf{1.0}} & \textcolor{mygray}{\textbf{72.6}}   & \textcolor{mygray}{\textbf{90.6}}  &  \underline{\textcolor{mygray}{93.4}}     &    \textcolor{mygray}{\textbf{2.0}}      & \textcolor{mygray}{\textbf{43.4}}  & \textcolor{mygray}{\textbf{70.7}}   & \textcolor{mygray}{\textbf{79.7}}  & \textcolor{mygray}{\textbf{2.0}} &    \textcolor{mygray}{\textbf{44.6}}    & \textcolor{mygray}{\textbf{71.2}}        &     \textcolor{mygray}{\textbf{79.7}}                  \\ 
\bottomrule
\end{tabular}
}
\caption{\textbf{Comparison with other work.} medR ($\downarrow$), Recall@k ($\uparrow$) are reported on the Recipe1M test set. Our approaches significantly outperform all existing work. Best metrics are in bold, and next best metrics are underlined.}
\label{tab:comparison}
\end{table*}

\paragraph{Multi Modal Transformer Decoder (MTD)}
Unimodal encoders encode each modality independently, and lack the ability to capture the interaction between the two modalities. Here we propose a multimodal transformer decoder (MTD) \cite{vaswani2017attention} that predicts the matching score of each pair of samples. We argue that this predicted matching score better captures  the interaction between the two modalities, compared to the global cosine similarity applied between the independently extracted embeddings. 
Specifically, MTD contains self-attention, cross-attention and feedforward modules. For the cross attention, the tokens of one modality are considered as queries (Q) and the tokens of the other modality as Keys and Values (KV). The matching score between a pair of samples can be obtained as follows:
\begin{equation}
    s(t_R, t_I) = MTD(t_R, \hat{t}_I) =  MTD(t_R, ITEM(t_I, t_R)),
\end{equation}
where $t_R$ and $\hat{t}_I$ are the output recipe tokens and the enhanced image tokens respectively.

\paragraph{Image-Text Matching (ITM) loss:}
ITM is a binary cross-entropy loss that is optimised to classify whether or not an image-text pair matches. Following the success of batch hard mining for triplet loss, the loss is applied on the hardest negative sample; for each recipe/image anchor we sample a negative recipe/image in the batch that is most similar to the anchor using the fast cosine similarity of the embeddings at the output of the DE. The loss can be written as follows:
\begin{align}
    \mathcal{L}_{itm} & = -\mathbb{E}_{t_R, t_I \sim D}[y\log(s(t_R, t_I)) + \\ \nonumber 
    & (1-y)\log(1-s(t_R, t_I))],
\end{align}
where $y$ is the label (\emph{i.e.}, 1 for matching pair and 0 otherwise) and $D$ is the set of pairs.

\subsection{Retrieval loss}

Following other work on cross-modal food retrieval \cite{Salvador_2017_CVPR_recipe1m, wang2019learning_acme,  salvador2021revamping, guerrero2021cross_xmrs}, we use the triplet loss in our learning framework. Here we introduce a new variant of triplet loss with dynamic margin.\\
\\
\textbf{Triplet loss:} The triplet loss can be written as:
 \begin{equation}
 \label{eq:triplet}
    l(x_a, x_p, x_n, \alpha) = [d(x_a, x_p) +\alpha - d(x_a, x_n)]_+,
\end{equation}
where $x_a$, $x_p$ and $x_n$ are the anchor, positive and negative examples respectively, $\alpha$ is the margin and $d(\cdot, \cdot)$ is a distance function.\\ 
\\
\textbf{IncMargin loss:} Inspired by the curriculum learning approaches \cite{curriculum}, where the task becomes harder as the training progresses, we propose to replace the constant margin by a dynamic one. The idea is that the difficulty of the task is also affected by the margin; a small margin is easier to optimize than a large margin. 
Here we assume that the task is difficult at the beginning, thus we make it simpler by starting with a small margin $\alpha_{inc}$ and increase it at each epoch until reaching a maximum value:
\begin{align}
  \mathcal{L}_{m}(\mathcal{B}_a, \mathcal{B}_p, \mathcal{B}_n) = & \sum_{x_a \in \mathcal{B}_a} \sum_{x_p \in \mathcal{B}_p} \sum_{x_n \in \mathcal{B}_n} l(x_a, x_p, x_n, \alpha_{inc})
\end{align}
We keep the adaptive margin between an acceptable minimum and maximum values.
In addition, we follow Adamine \cite{carvalho2018adamine} and use the triplet loss with an adaptive weighting strategy, where the triplet is weighted by a similar term as $\delta$ to overcome the vanishing update when most of the triplets are inactive. We call the instance loss as image-text contrastive loss which can be written as: 
\begin{align}
 \mathcal{L}_{itc} = & \frac{1}{\delta_{r}} \mathcal{L}_{m}(\mathcal{B}_a^r, \mathcal{B}_p^v, \mathcal{B}_n^v) + \frac{1}{\delta_{v}} \mathcal{L}_{m}(\mathcal{B}_a^v, \mathcal{B}_p^r, \mathcal{B}_n^r) 
\end{align}
Where $\mathcal{B}^r$ and $\mathcal{B}^v$ are the sets of recipe and image embeddings, $\delta_{r}$ and $\delta_{v}$ are the number of triplets that contribute to the loss ($l > 0$ Eq. \ref{eq:triplet}) when the recipe and image embeddings are anchors respectively. The positive and negative samples are from the other modality; the positive is the one associated with the anchor and other samples in the batch are considered negatives.

Similarly, we incorporate the semantic triplet loss of Adamine \cite{carvalho2018adamine} as a regularization to obtain more semantically rich embeddings. The semantic loss $\mathcal{L}_{sem}$ is the same as the instance loss except for the selection of positive and negative samples. Here the positive samples are those that share the same class with the anchor and the negative samples are the samples with different class.\\
\\
\textbf{Total loss:}
Thus, the total loss can be written as follows:
\begin{equation}
    \mathcal{L} = \mathcal{L}_{itc} + \lambda_{sem}\mathcal{L}_{sem} + \lambda_{itm}\mathcal{L}_{itm}
\end{equation}
Where $\lambda_{sem}$ and $\lambda_{itm}$ are the corresponding losses weights.

\section{Experiments} 
\label{sec:exp}
We use Recipe1M dataset \cite{Salvador_2017_CVPR_recipe1m} with 238,999, 51,119, 51,303 pairs as training, validation and test set. We report median rank (medR) the median of the retrieved samples index for each query, and recall@K (\emph{i.e.}, R@1, R@5, R@10) the percentage of queries for which the correct sample index belongs to the top K retrieved samples. We report the mean over 10 (resp.5) bags of 1k (resp. 10k) pairs of the Recipe1M test set.\\
\setlength{\tabcolsep}{5pt}
\begin{table*}[h]
\centering
\begin{tabular}{cccccc|cccc|cccc@{}}
\toprule
 \multicolumn{1}{c|}{} & \multicolumn{1}{c|}{RE} & \multicolumn{2}{c|}{MMR} & \multicolumn{1}{c|}{$\mathcal{L}_{m}$} & \multicolumn{1}{c|}{CLIP} & \multicolumn{4}{c|}{\textbf{image-to-recipe}} & \multicolumn{4}{c}{\textbf{recipe-to-image}} \\ \midrule 
B  & \multicolumn{1}{|c}{HTD} & \multicolumn{1}{|c}{MTD} & \multicolumn{1}{c|}{ITEM} & \multicolumn{1}{c|}{IncMargin} & \multicolumn{1}{c|}{ViT-CLIP}                  & medR     & R@1      & R@5    & R@10    & medR     & R@1     & R@5     & R@10    \\ \midrule
\greencheck & & &  & &    &  1 &59.5 &83.1 &88.1 &1 &61.1 &83.5 &88.2 \\ 
\greencheck & \greencheck & &  & &  &  1 &65.6 &86.7 &90.8 &1 &66.8 &86.9 &90.8 \\
\greencheck & \greencheck &\greencheck &    & & & 1 & 66.9  & 87.0  &91.0  &1 &67.5  &87.2  &91.1             \\ 
\greencheck & \greencheck &\greencheck &\greencheck  & &  & 1 & 67.0  & 87.5  &90.9  &1 &68.1  &87.4  &91.0           \\
\greencheck & \greencheck &\greencheck &\greencheck  & \greencheck &   & 1 & 68.2  & 87.9  &91.3  &1 &68.3  &87.8  & 91.5           \\
\greencheck & \greencheck & \greencheck & \greencheck  & \greencheck & \greencheck  &  \textbf{1}   &  \textbf{72.3}  & \textbf{90.7}    & \textbf{93.4}  &\textbf{1} & \textbf{72.6}   & \textbf{90.6}  &  \textbf{93.4}            \\
\bottomrule
\end{tabular}
\caption{Ablation study. medR ($\downarrow$), Recall@k ($\uparrow$) are reported on the Recipe1M test set with 1k setup. Each modification brings additional improvement as well as the combination of all of them. }
\label{tab:ablation}
\end{table*}
\\
\textbf{Model details:} We use ViT-B/16 and CLIP-ViT-B/16  as image encoders. For the recipe encoder, similarly to \cite{salvador2021revamping}, we use transformer encoders with 2 layers and 4 heads for hierarchical transformers $T$ and $HT$. For $HTD$ we use a transformer decoder (wihout masking) with 2 layers and 4 heads. The hidden layer dimension is kept 512 in the recipe encoder. The image and recipe embeddings are obtained with a different linear layers of output dimension 1024. The image and recipe tokens are projected using different linear layers of the same output dimension 1024 before going to the $MMR$ module. To keep the model simpler, the $ITEM$ module consists of a transformer decoder of only 1 layer and 4 heads with hidden size 1024. The $MTD$ consists of a transformer decoder with 4 layers, 4 heads and hidden dimension 1024. All the transformers are trained from scratch, except the ViT, which is initialised by ImageNet weights, and CLIP-ViT with CLIP weights. \\
\\
\textbf{Training details:} We mainly follow the implementation details of Adamine \cite{carvalho2018adamine}. Specifically, For the alignment loss we use $\lambda_{sem}=0.1$ for the semantic triplet loss and $\lambda_{itm}=1$. For each batch, half of the samples are associated with a class for the semantic loss. To compute the triplet loss, each sample is considered as anchor, the associated sample from the other modality as positive sample and all the others samples from the other modality in the batch as negatives. For the semantic triplet loss, the samples with the same class are considered positive samples and negative samples otherwise. The image encoder is kept frozen for the first 20 epochs, then all the modules are trained with a constant learning rate $1e-5$ (except CLIP-ViT with learning rate $1e-6$) and Adam optimiser. The models are trained until reaching 120 epochs with batch size 100. The models are trained with 2 A100 GPUs 40 GB. For IncMargin, we start by a $\alpha_{inc}=0.05$ and increase it by $0.005$ each epoch until reaching $0.3$. We use a margin $\alpha=0.3$ for all other experiments.

\subsection{Comparison with Other Work}
We compare our work to existing approaches in Table \ref{tab:comparison}. For Adamine, we shifted the values of medR as they assume medR=0 is for exact match, while other approaches adopt medR=1, we also notice that their metrics for recipe to image and image 2 recipe should be swapped. For H-T \cite{salvador2021revamping}, we use their official code to re compute the metrics with the ViT image encoder (H-T (ViT)) and we report these metrics for fair comparison with our approach.

We can notice that our approach significantly outperforms all other ones by a large margin for medR and R@1 with the 1k setup and for all the metrics with the 10k setup on the Recipe1M test set. 

Compared to the 2 SOTA H-T \cite{salvador2021revamping} and X-MRS \cite{guerrero2021cross_xmrs}, we surpass them on the more challenging 10k setup in terms of R@k (+ 6.5 \% and + 7.1 \% R@1/10k resp.)  and medR, and on the smaller setup 1k in terms of R@1 (+3.8 \% and + 4.3 \% R@1/1k resp.). We are largely better than the adversarial based approaches; +15.5 \% R@1/1k compared to ACME \cite{wang2019learning_acme} and +27.7 \%  R@1/1k compared to  \cite{Zhu_2019_CVPR_r2gan}. The same hold for the methods that use cross attention modules between the 2 modalities \cite{multisubspace} (+8.5\% R@1/1k).

Interestingly, our approach with CLIP-ViT outperforms other approaches by a large margin (+8.1 \% R@1/1k and +9.9 \% R@1/10k compared to H-T).

\begin{figure}
    \centering
\begin{tikzpicture}
\pgfplotsset{
        compat=1.7,
        my ybar legend/.style={
            legend image code/.code={
                \draw [##1] (0cm,-0.6ex) rectangle +(2em,1.5ex);
            },
        },
    }
\begin{axis}[
	symbolic x coords={10k,20k,30k,40k,50k},
    xtick=data,  
    xticklabel style={text height=2ex},
	ylabel=medR,
	enlargelimits=0.15,
	legend style={at={(0,1)},
	anchor=north west,legend columns=1, font=\tiny},
	width=\linewidth,
	ybar = 0,
	bar width=5.5pt,
	nodes near coords,
	every node near coord/.append style={xshift=-10pt,yshift=8pt,anchor=north west,font=\tiny},
	legend image code/.code={
        \draw [#1] (0cm,-0.1cm) rectangle (0.1cm,0.12cm); }
]
\addplot [draw=pink, fill=pink] 
	coordinates {(10k,2) (20k,3)
		 (30k,4) (40k, 5) (50k, 6)};
\addplot [draw=red, fill=red] 
	coordinates {(10k,2) (20k,4)
		 (30k,5) (40k, 6) (50k, 8)};
\addplot [draw=orange, fill=orange] 
	coordinates {(10k,3) (20k,5)
		 (30k,7) (40k, 9) (50k, 11)};
\addplot [draw=yellow, fill=yellow] 
	coordinates {(10k,4.2) (20k, 8.0)
		 (30k, 11.9) (40k, 15.0) (50k, 19.0)};
\addplot [draw=green, fill=green] 
	coordinates {(10k,5) (20k,9.4)
		 (30k,13.9) (40k, 18.6) (50k, 23.8)};
\addplot [draw=blue, fill=blue] 
	coordinates {(10k, 6.2) (20k, 12.7)
		 (30k, 18.9) (40k, 23.5) (50k, 31.2)};
\legend{T-Food (CLIP-ViT),
T-Food (ViT), 
H-T (ViT) \cite{salvador2021revamping}, 
SEJE \cite{xie2021learning2}, 
HF-ICMA \cite{hybrid_fusion}, 
IMHF \cite{intra_inter}}
\end{axis}
\end{tikzpicture}
    \caption{Scalability analysis: we report the medR ($\downarrow$) of image-to-recipe retrieval on Recipe1M test set with setup more than 10k.}
    \label{fig:scalability}
\end{figure}
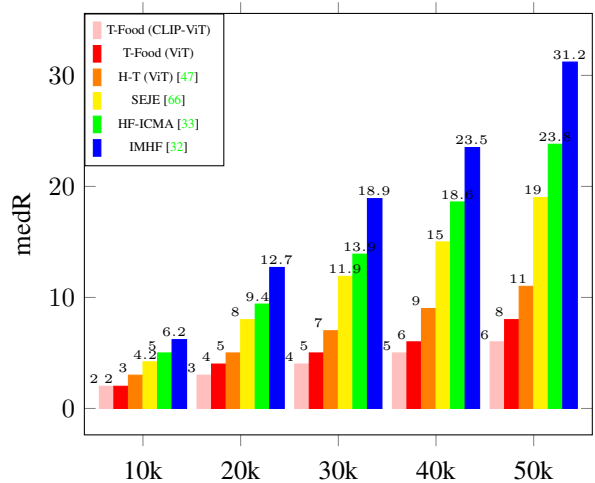

In Figure \ref{fig:scalability}, we test our approach on different test sizes more than 10k, we can notice that the performance gap increases further between our and other approaches.

\setlength{\tabcolsep}{2.5pt}
\begin{table}[th]
\small
\centering
\begin{tabular}{@{}l|ccc|ccc@{}}
\toprule
 & \multicolumn{3}{c|}{\textbf{image-to-recipe}} & \multicolumn{3}{c}{\textbf{recipe-to-image}} \\ \midrule 
            Margin scheme & \small{R@1}      & \small{R@5}    & \small{R@10}        & \small{R@1}     & \small{R@5}     & \small{R@10}    \\ \midrule
Baseline B  &  59.5  &  83.1  &88.1  &61.1  &83.5  & 88.2           \\ 
\midrule
Fixed $\alpha$  & 59.5  & 83.4  &88.1  &60.7  &83.5  & 88.3             \\ 
\small{AdaMargin $\alpha/\delta$} &  60.8  &  83.4  &88.2  &61.5  &83.7  &88.4           \\
\small{IncMargin $\alpha_{inc}$}& \textbf{62.9}  & \textbf{84.8}  &\textbf{89.4}   &\textbf{63.8}  &\textbf{85.1}  &\textbf{89.3}           \\ \bottomrule
\end{tabular}
\caption{Ablation study for AdaMargin and IncMargin on the Recipe1M test set with 1k setup. We compare our baseline model B (Sec. \ref{sec:ablation}) to three variants trained without Adamine. Our new triplet loss IncMargin outperforms significantly AdaMargin and the classical triplet loss with fixed margin. Both AdaMargin and IncMargin outperforms B.}
\label{tab:ablation_triplet}
\end{table}

\subsection{Ablation Study}
\label{sec:ablation}
\begin{figure}
\centering
\begin{tikzpicture}
\begin{axis}[%
scatter/classes={%
    a={mark=*,draw=black}},
    xlabel=$R@1$,
    ylabel=$R@1$ with re-ranking,
    legend pos=south east,
    xmin=52, xmax = 75,
    ymin=52, ymax= 75,
  ]
    \draw [dotted] (55, 52) -- (55, 75);
    \draw [dotted] (60, 52) -- (60, 75);
    \draw [dotted] (65, 52) -- (65, 75);
    \draw [dotted] (70, 52) -- (70, 75);
    \draw [dotted] (52, 55) -- (75, 55);
    \draw [dotted] (52, 60) -- (75, 60);
    \draw [dotted] (52, 65) -- (75, 65);
    \draw [dotted] (52, 70) -- (75, 70);
  
\addplot[dashed, domain=50:80, samples=100, color=black, mark=]
{x};
\addlegendentry{\tiny{line $y=x$}}
\addplot[only marks, mark=*, color=violet]
coordinates{
  (54.0, 63.6)
}; 
\addlegendentry{\tiny{MTD w/o Adamine}}

\addplot[only marks, mark=*, color=red]
coordinates{
  (58.8, 64.2)
}; 
\addlegendentry{\tiny{MTD}}

\addplot[only marks, mark=*, color=orange]
coordinates{
  (66.9, 66.6)
}; 
\addlegendentry{\tiny{HTD + MTD}}

\addplot[only marks, mark=*, color=green]
coordinates{
  (67.0, 67.5)
}; 
\addlegendentry{\tiny{HTD + MTD + ITEM}}

\addplot[only marks, mark=*, color=cyan]
coordinates{
  (68.2, 68.2)
}; 
\addlegendentry{\tiny{T-Food}}

\addplot[only marks, mark=*, color=blue]
coordinates{
  (72.3, 72.2)
}; 
\addlegendentry{\tiny{T-Food (CLIP)}}

\end{axis}
\end{tikzpicture}
\caption{Image-to-Recipe R@1 with and without multimodal re-ranking (the top 10 examples for each query are re-ranked using MTD) for different variants of our approach. Both metrics converge with our best models.}
\label{fig:reranking}
\end{figure}
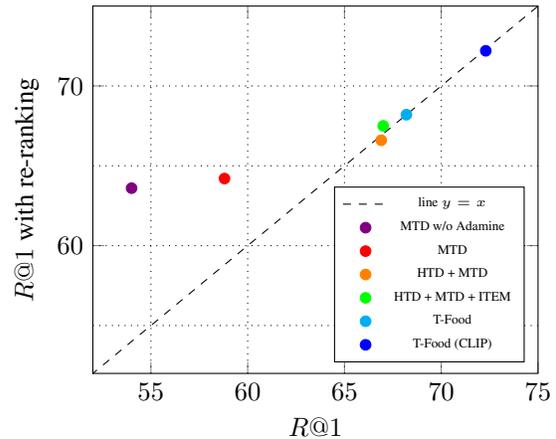

In this section, we investigate the importance of different design choices. Our baseline (B) consists of the dual encoders; ViT as image encoder, hiererchical transformer as recipe encoder (\emph{i.e.}, with T and HT), and Adamine loss.

From Table \ref{tab:ablation} we can notice that the proposed \textbf{HTD} leads to better alignment, seen by the significant improvement of metrics, this validates our hypothesis that the recipe elements are highly entangled and should not be embedded separately. 
We can also notice that \textbf{MTD} also brings additional improvement, this is consistent with the findings of VLP in general and reveals that multimodal transformers are promising for the underlying task. \\
\textbf{ITEM} brings additional improvements, which in some sense facilitates the role of MTD, by fusing further the image with the recipe tokens. Using \textbf{ViT-CLIP} as image encoder brings significant improvement, this indicates that using models trained on large datasets can cope with the noise in this relatively small dataset.

From table \ref{tab:ablation_triplet} and \ref{tab:ablation}, our \textbf{IncMargin} loss, based on an increasing margin value $\alpha_{inc}$, improves upon the classical triplet loss by more than 3 points (image to recipe R@1). In addition, we compare to AdaMargin, another variant of triplet loss with adaptive margin that changes according to the number of active triplet in the batch (see supplementary material). Despite its simplicity, IncMargin also outperforms AdaMargin. From our experiments, we noticed that the value of the margin increases to its maximum value faster with AdaMargin, while the evolution of the margin is smoother and takes longer time to each its maximum value with IncMargin. In addition, we can notice that the improvement of this variant are smaller with the ITM loss (from 67.0 to 68.2 R@1 image-to-recipe) , this should be expected as the model focuses on the ITM regularization when the triplet loss is small (\emph{i.e.}, the margin is small at the beginning of the training) and we noticed from our experiments that the triplet/retrieval loss is more important than the ITM for retrieval.\\

\setlength{\tabcolsep}{2.5pt}
\begin{table}[th]
\small
\centering
\begin{tabular}{@{}l|ccc|ccc@{}}
\toprule
 & \multicolumn{3}{c|}{\textbf{image-to-recipe}} & \multicolumn{3}{c}{\textbf{recipe-to-image}} \\ \midrule 
       nb. of layers              & \small{R@1}      & \small{R@5}    & \small{R@10}       & \small{R@1}     & \small{R@5}     & \small{R@10}    \\ \midrule
\small{w/o MMR} &65.6 &86.7 &90.8 &66.8 &86.9 &90.8  \\ 
\multirow{1}{*}{\small{1}} & 66.8  & 87.0  &90.4  &67.0  &86.9  & 90.4  \\ 
\multirow{1}{*}{\small{4}} & 67.5  & 87.5  &91.0  &67.8  &87.6  & 91.3  \\ 
\multirow{1}{*}{\small{6}} & 65.7  & 86.6  &90.3  &66.1  &86.5  & 90.3  \\ 

\bottomrule
\end{tabular}
\caption{Ablation study for MMR on the test set with 1k setup. We report the performance w.r.t the number of layers in MTD. All the models are trained with B + HTD + MMR and $\lambda_{itm} =0.1$ .}
\label{tab:ablation_MR_mtd}
\end{table}
\textbf{MMR:} 
To perform image/recipe retrieval, we compute the pairwise cosine similarity of each pair of image/recipe in the shared embedding space, which can be done efficiently in parallel. Other work \cite{miech2021thinking_fastslow, li2021alignalbef} propose to use multimodal modules to re-rank the top k retrieval propositions for each query. Since these multimodal modules compute a more fine-grained image/text alignment, it improves the retrieval metrics, but with a much higher computational cost. In Figure \ref{fig:reranking}, we compare those two retrieval methods for our model variants that use a MMR module. We can see that our model MTD w/o Adamine has a low R@1 score without re-ranking (54.0) but much higher with multimodal re-ranking (63.6), emphasizing the role of multimodal alignment in that case. Adamine helps to partially close this gap by giving more importance to the shared embedding space, which improves greatly the retrieval metrics without re-ranking, with small improvements of the metrics with re-ranking. When we add the HTD module (for recipe encoding), we observe that the role of re-ranking becomes useless. This remains true for our best models T-Food and T-Food (CLIP). Therefore, we conclude that the improvement brought about by re-ranking was due to the lack of expressiveness of the recipe encoder and not the need for a finer image/text alignment. We also conclude that the presumably better alignment capabilities of the MMR module are fully transferred to the shared embedding space at the end of training, given that re-ranking with this module is not better than global embedding similarity.
We also investigated how the complexity of MMR affects the performance in Table \ref{tab:ablation_MR_mtd}, where we train our model with different number of layers in MTD. We can notice that when MTD becomes more powerful, the metrics become better (nb. of layers = 1 vs 4).  Note that, these results are consistent with other contrastive learning work, specifically, in SimCLR \cite{chen2020simple_simclr}, the authors show better results when the projection head is more powerful (\emph{e.g.}, nonlinear instead of linear), and here we make the analogy between their projection head and our MRR module. However, a too heavy MTD degrades the results (nb. of layers = 4 vs 6), we argue that when the regularization module becomes too powerful, most of the modality fusion and cross modal interaction will be focused in it, instead of the DE.

We can notice that each modification brings additional improvement as well as the combination of them.

\vspace*{-.15cm}{\section{Conclusion}
\label{sec:conclusion}
\vspace*{-.1cm}We proposed a new framework for multimodal alignment in the cooking context and validated the approach by surpassing all existing work by a large margin on the cross modal retrieval task. 
We introduced a novel MultiModal Regularization and new variant of triplet loss, in addition we showed that transformer encoders/decoders can bring significant improvement by exploiting the intra and inter dependencies between modalities and recipe elements. Finally, we showed that transfer learning from VLP models trained on large datasets can significantly help in this context. While the work is successful in the food domain, we believe that the proposed ideas can be beneficial for cross modal tasks in general, especially cross modal retrieval.}\\

\vspace*{0.cm}{\noindent\textbf{Acknowledgments}: This work was partly supported by ANR grant VISA DEEP (ANR-20-CHIA-0022), and HPC resources of IDRIS under the allocation 2021-[AD011013159] made by GENCI.}

\appendix
\section{Appendix}

\subsection{AdaMargin Triplet Loss}
Here we detail another variant that we propose for adaptive triplet with dynamic margin. 
The number of active triplets (the triplets corresponding to non-zero triplet loss $l$) in the batch reveals the difficulty of the task; if it is small, that means most of the triplets already satisfy the triplet condition (\emph{i.e.}, the difference between the distance of positive pairs is smaller than the distance of the negative pairs by a margin). Thus, we propose to inversely weigh the margin by the number of active triplets $\delta$ in each batch. Thus the loss can be written as:
\begin{align}
 \mathcal{L}_{m}(\mathcal{B}_a, \mathcal{B}_p, \mathcal{B}_n) = & \sum_{x_a \in \mathcal{B}_a} \sum_{x_p \in \mathcal{B}_p} \sum_{x_n \in \mathcal{B}_n} l(x_a, x_p, x_n, \alpha/\delta)  
\end{align}
Where $\mathcal{B}_a$, $\mathcal{B}_p$ and $\mathcal{B}_n$ are the set of anchors, positive and negative examples in the batch.
We keep the $\alpha/\delta$ between $0.05$ and $0.3$, and $\delta$ is computed based on $\alpha = 0.3$

\subsection{Ablation Study}
\label{sec:app_ablation}

In this section, we investigate the importance of different design choices. Our baseline (B) consists of the dual encoders, ViT for image encoder, hiererchical transformer (HT) for recipe encoder (\emph{i.e.}, with T and HT), and trained with Adamine loss.

\paragraph{HTD:} In Table \ref{tab:ablation_htd}, we investigate the importance of the title for the instructions and ingredients, thus we compare our approach with another variant that takes only the ingredients (resp. instructions) as K and V with the instructions (resp. ingredients) as queries. We can notice that the title does not bring additional improvement to the instructions and ingredients.
\setlength{\tabcolsep}{2.5pt}
\begin{table}[th]
\small
\centering
\begin{tabular}{@{}l|cccc|cccc@{}}
\toprule
\multirow{2}{*}{} & \multicolumn{4}{c|}{\textbf{image-to-recipe}} & \multicolumn{4}{c}{\textbf{recipe-to-image}} \\ \cmidrule(l){2-9} 
                  & \small{medR}     & \small{R1}      & \small{R5}    & \small{R10}    & \small{medR}     & \small{R1}     & \small{R5}     & \small{R10}    \\ \midrule
\small{Ours}    & 1 & 66.9  & 87.0  &91.0  &1 &67.5  &87.2  & 91.1             \\ 
\small{v2}   & \textbf{1} & \textbf{67.3}  & \textbf{87.3}  &\textbf{91.1}  &\textbf{1} &\textbf{67.6}  &\textbf{87.5}  &\textbf{91.2}           \\ \bottomrule
\end{tabular}
\caption{Ablation study for HTD. medR ($\downarrow$), Recall@k ($\uparrow$) are reported on the Recipe1M test set with 1k setup. v2: in HTD, the cross attention of ingredients takes only the instructions as K and V and the one for instructions takes only the ingredients as K and V, while the title takes both of them as K and V.}
\label{tab:ablation_htd}
\end{table}
\paragraph{ITEM:} We did an ablation study for ITEM in Table \ref{tab:ablation_item}. In particular, we test the module with all recipe elements as K and V (ITEM (a)), with the title alone (ITEM (t)) and with ingredients alone (ITEM (n)). The best results are obtained with all recipe elements, which also validate that all recipe elements are important for multimodal fusion.\\

\setlength{\tabcolsep}{2.5pt}
\begin{table}[th]
\small
\centering
\begin{tabular}{@{}l|cccc|cccc@{}}
\toprule
\multirow{2}{*}{} & \multicolumn{4}{c|}{\textbf{image-to-recipe}} & \multicolumn{4}{c}{\textbf{recipe-to-image}} \\ \cmidrule(l){2-9} 
                  & \small{medR}     & \small{R1}      & \small{R5}    & \small{R10}    & \small{medR}     & \small{R1}     & \small{R5}     & \small{R10}    \\ \midrule
\small{B*}    & 1 & 66.6  & 87.5  &91.0  &1 &67.6  &88.1  &91.1             \\ 
\small{B* + ITEM (t)}   & 1 &  65.8  &  87.9  & \textbf{91.4}  &1 &66.0  &88.4  & \textbf{91.4}           \\ 
\small{B* + ITEM (n)}   & 1 &  66.7  &  87.9  &91.1  &1 &67.2  &88.4  &91.1           \\
\small{B* + ITEM (a)}   & \textbf{1} & \textbf{67.5}  & \textbf{88.1}  &90.9  &\textbf{1} &\textbf{68.4}  &\textbf{88.6}  & 91.0           \\ \bottomrule
\end{tabular}
\caption{Ablation study. medR ($\downarrow$), Recall@k ($\uparrow$) are reported on the Recipe1M test set with 1k setup. B* is with ViT, HT, HTD and MTD. We compare several ways of enhancing the image tokens, the cross attention is with; all recipe elements (ITEM (a)), only the title (ITEM(t) and only the ingredients (ITEM (n)).  The cross attention between the image and all recipe elements gives the best results.}
\label{tab:ablation_item}
\end{table}%

{\small
\bibliographystyle{ieee_fullname}
\bibliography{tfood_arxiv}
}

\end{document}